\documentclass[11pt,a4paper]{article}
\usepackage[hyperref]{emnlp2020}
\usepackage{times}
\usepackage{latexsym}

\usepackage{microtype}

\aclfinalcopy %
\def\aclpaperid{1409} %

\usepackage{amsfonts}
\usepackage{bm}
\usepackage{array,amsmath}
\usepackage{amssymb}
\usepackage{dsfont}
\usepackage{float}
\usepackage{graphicx}
\usepackage{algorithm}
\usepackage{algorithmicx}
\usepackage{algpseudocode}
\usepackage{pgf,tikz}
\usepackage{mathrsfs}
\usepackage{multirow}
\usepackage{booktabs}
\usetikzlibrary{arrows}
\usepackage{threeparttable}

\def\figref#1{Figure~\ref{fig:#1}}
\def\figlabel#1{\label{fig:#1}\label{p:#1}}

\def\tabref#1{Table~\ref{tab:#1}}
\def\tablabel#1{\label{tab:#1}\label{p:#1}}

\def\secref#1{\S\ref{sec:#1}}
\def\seclabel#1{\label{sec:#1}}
\def\eqref#1{Eq.~\ref{eqn:#1}}

\def\numpar{100k}
\def\ppnumpar{5}

\newcounter{notecounter}
\newcommand{\enotesoff}{\long\gdef\enote##1##2{}}
\newcommand{\enoteson}{\long\gdef\enote##1##2{{
			\stepcounter{notecounter}
			{\large\bf
				\hspace{1cm}\arabic{notecounter} $<<<$ ##1: ##2
				$>>>$\hspace{1cm}}}}}
\enoteson
\enotesoff
\long\def\eat#1{}

\def\dnrm#1{\mbox{$_{\hbox{\scriptsize #1}}$}}

\title{SimAlign: High Quality Word Alignments
  Without Parallel Training Data Using Static and Contextualized Embeddings}

\author{Masoud Jalili Sabet\thanks{\mbox{\ \ } Equal contribution - random order.}$\ \,^1$, Philipp Dufter$^{*}$$^1$, Fran\c{c}ois Yvon$^2$, Hinrich Sch\"{u}tze$^1$\\
$^1$ Center for Information and Language Processing (CIS), LMU Munich, Germany\\
$^2$ Universit\'{e} Paris-Saclay, CNRS, LIMSI, France\\
{\tt \{masoud,philipp\}@cis.lmu.de,francois.yvon@limsi.fr}}

\date{}

\begin{document}
\maketitle 

\begin{abstract} 
	Word alignments are useful for tasks like statistical and
	neural machine translation (NMT) and cross-lingual annotation projection.
	Statistical word aligners 
	perform well, as do methods that extract
	alignments jointly with translations in NMT.
	However, most approaches require
	parallel training data, and quality decreases as less
	training data is available.  We propose word
	alignment methods that require no parallel
	data. The key idea is to leverage multilingual word
		embeddings -- both static and contextualized -- for word alignment. Our
	multilingual embeddings are created from monolingual data
	only without relying on any parallel data or dictionaries.
	We find that alignments created from embeddings
	are superior for four and comparable for two language pairs compared to those produced by 
	traditional statistical aligners 
        -- even with abundant
	parallel data; e.g.,
contextualized embeddings achieve
	a word alignment $F_1$ for English-German that is \ppnumpar \ percentage points higher than eflomal,
	a high-quality
	statistical aligner,
	trained on \numpar \ parallel sentences.
\end{abstract}

\section{Introduction}
Word alignments are essential for statistical machine translation and 
useful in NMT, e.g., for imposing  priors on 
attention matrices \cite{liu2016neural,chen2016guided,alkhouli2017biasing,alkhouli2018alignment} 
or for  decoding \cite{alkhouli2016alignment,press2018you}. Further, word alignments have been
successfully used in a  range of tasks such as
typological analysis \cite{lewis2008automatically,ostling2015word},
annotation projection \cite{yarowsky2001inducing,pado2009cross,asgari2017past,huck2019cross} and 
creating multilingual embeddings \cite{guo2016representation,ammar2016massively,dufter-18-embedding}.

Statistical word aligners such as the IBM models \cite{brown1993mathematics} and their 
implementations 
Giza++ \cite{och03:asc}, fast-align \cite{dyer2013simple}, as well as newer models such 
as eflomal \cite{ostling2016efficient}
are widely used for alignment.
With the rise of NMT
\cite{bahdanau2014neural},
attempts have been made to interpret attention matrices as soft word alignments \cite{cohn-etal-2016-incorporating,koehn2017six,ghader2017does}. 
Several methods create alignments from attention matrices \cite{peter2017generating,zenkel2019adding}
or pursue a multitask approach for alignment and translation \cite{garg2019jointly}. However, 
most systems require parallel data (in sufficient amount to train high quality NMT systems) and their performance 
deteriorates when  parallel text is scarce (Tables 1--2 in \citep{och03:asc}).

\begin{figure}
	\centering
	\begin{tabular}{c}
	\includegraphics[width=0.93\linewidth]{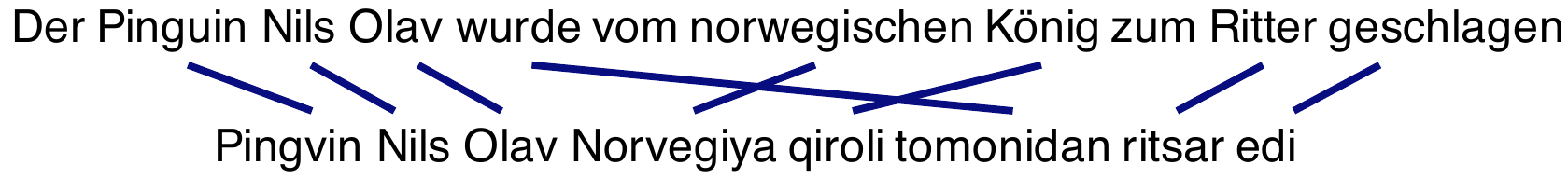}\\
\midrule
\includegraphics[width=0.93\linewidth]{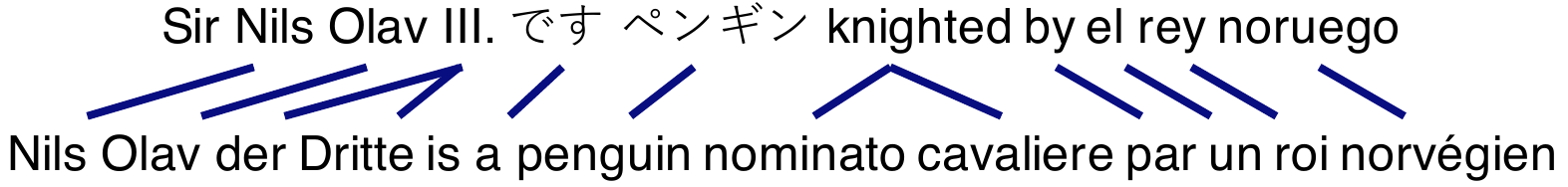}
	\end{tabular}
	\caption{Our method does not rely on parallel
          training data and
          can align 
		distant language pairs (German-Uzbek, top) and even 
	mixed sentences (bottom). Example sentence is manually created. Algorithm: Itermax.}
	\figlabel{firstpageexample}
\end{figure}

Recent unsupervised multilingual embedding algorithms that use only non-parallel data
provide high quality static \cite{artetxe2018robust,conneau2018word} and contextualized embeddings
\cite{devlin2019bert,conneau2019unsupervised}.
\emph{Our key idea is to leverage these embeddings
  for word alignments
-- by 
extracting alignments from similarity matrices induced from
embeddings -- without relying on  parallel data.} Requiring no or little
parallel data is advantageous, e.g., in
the low-resource case 
and in domain-specific settings without parallel data.
A lack of parallel data cannot be easily remedied:
mining parallel sentences is possible
\cite{schwenk2019wikimatrix} but
assumes that
comparable, monolingual corpora contain
parallel sentences. Further, we find that large amounts of mined parallel data 
do not necessarily improve alignment quality.

Our main
\textbf{contribution} is that
we show that \emph{word alignments obtained
  from multilingual pretrained language models
  are superior for four and comparable for two language pairs, compared to strong statistical word
  aligners like eflomal 
even in high resource scenarios.} Additionally,
\textbf{(1)} we
introduce three new alignment methods based on the matrix of
embedding similarities and two extensions that handle null words and integrate positional information.
They permit a flexible tradeoff of recall and precision.
\textbf{(2)} We provide evidence that subword processing is beneficial for aligning rare words.
\textbf{(3)} We bundle the source code of our methods in a tool called \emph{SimAlign}, which is available.\footnote{\url{https://github.com/cisnlp/simalign}}
An interactive online demo is available.\footnote{\url{https://simalign.cis.lmu.de/}}

\section{Methods}

\subsection{Alignments from Similarity Matrices}

We propose three methods to obtain alignments from similarity matrices. 
Argmax is a simple baseline, IterMax a novel iterative algorithm, 
and Match a graph-theoretical method based on identifying matchings in a bipartite graph.

Consider parallel sentences $s^{(e)},s^{(f)}$,
with lengths $l_e, l_f$
in 
languages $e,f$.
Assume we have access to some embedding function $\mathcal{E}$ that
maps each word in a sentence to a $d$-dimensional vector, i.e., $\mathcal{E}(s^{(k)}) \in \mathbb{R}^{l_k \times d}$
for $k \in \{e,f\}$. Let $\mathcal{E}(s^{(k)})_i$ denote the vector of the $i$-th word in  sentence $s^{(k)}$. For static embeddings $\mathcal{E}(s^{(k)})_i$ depends only on the word $i$ in language $k$ whereas for contextualized embeddings the vector depends on the full context $s^{(k)}$. We define the \emph{similarity 
	matrix} as the matrix $S \in [0,1]^{l_e \times l_f}$ induced by
the embeddings
where $S_{ij} := \text{sim}\left(\mathcal{E}(s^{(e)})_i, \mathcal{E}(s^{(f)})_j\right)$
is some normalized measure of similarity, e.g., cosine-similarity normalized 
to be between $0$ and $1$. 
We now describe our methods for extracting alignments
from $S$, i.e., obtaining a binary matrix $A \in \{0, 1\}^{l_e \times l_f}$.

\textbf{Argmax.}
A simple baseline is to align $i$ and $j$ when $s^{(e)}_i$ is the most similar word to $s^{(f)}_j$ and vice-versa.
That is, we set $A_{ij} = 1$ if
\begin{align*}
(i = \arg\max_l S_{l,j}) \wedge  (j = \arg\max_l S_{i,l})
\end{align*}
and $A_{ij} = 0$ otherwise. In case of ties, which are unlikely in similarity matrices, we 
choose the smaller index. 
If all entries in a row $i$ or column $j$ of $S$ are $0$ we set $A_{ij} = 0$ (this case can appear in Itermax).
Similar methods have been applied to co-occurrences \cite{melamed-2000-models} (``competitive linking''),
Dice coefficients \cite{och03:asc} and attention matrices \cite{garg2019jointly}.

\textbf{Itermax.}
There are many sentences for which 
Argmax only identifies few alignment edges 
because mutual argmaxes can be rare.
As a remedy, we apply Argmax iteratively. Specifically, we modify
the similarity matrix conditioned on the alignment edges found in a previous iteration: if two words $i$ and $j$
have \emph{both} been aligned, we zero out the similarity. Similarly, if \emph{neither} is aligned we leave the 
similarity unchanged. In case only one of them is aligned, we multiply the similarity with a discount factor $\alpha\in[0,1]$.
Intuitively, this encourages the model to focus on unaligned word pairs. However, if the similarity with an already aligned
word is exceptionally high, the model can add an additional edge. 
Note that this explicitly allows one token to be aligned to multiple other tokens. For details on the algorithm see  \figref{itermax}.

\begin{figure}
	\begin{algorithm}[H]
		\small
		\algrenewcommand\algorithmicindent{0.3cm}
		\caption{Itermax.}
		\begin{algorithmic}[1]
			\Procedure{Itermax}{$S$, $n_{\dnrm{max}}$, $\alpha \in [0,1]$}
			\State $A, M = \text{zeros\_like}(S)$%
			\For {$n \in [1,\ldots, n_{\dnrm{max}}]$}
			\State $\forall i,j:$
			\State $M_{ij} = \begin{cases}
			1 \text{ if }\max\left(\sum_{l=0}^{l_e}A_{lj}, \sum_{l=0}^{l_f}A_{il}\right) = 0\\
			0 \text{ if }\min\left(\sum_{l=0}^{l_e}A_{lj}, \sum_{l=0}^{l_f}A_{il}\right) > 0\\
			\alpha \text{ otherwise}
			\end{cases}$
			\State $A_{\dnrm{to\_add}} = \text{get\_argmax\_alignments}(S \odot M)$
			\State $A = A + A_{\dnrm{to\_add}}$
			\EndFor
			\State \Return $A$
			\EndProcedure
		\end{algorithmic}
	\end{algorithm}
	\caption{Description of the Itermax algorithm.  
		\emph{zeros\_like} yields a matrix with zeros and with same shape as the input, 
		\emph{get\_argmax\_alignments} returns alignments obtained using the Argmax Method,
		$\odot$ is elementwise multiplication. \figlabel{itermax}}
\end{figure}

\textbf{Match.}
Argmax finds a local, not a global optimum and Itermax is a greedy algorithm.
To find global optima,
we  frame  alignment as an assignment
problem: we search for a maximum-weight maximal
matching (e.g., \cite{kuhn1955hungarian}) in the bipartite weighted graph 
which is induced by
the similarity matrix.
This optimization problem is defined by 
\begin{equation*}
A^* = \textrm{argmax}_{A \in \{0, 1\}^{l_e \times l_f}} \sum_{i=1}^{l_e}\sum_{j=1}^{l_f}A_{ij}S_{ij}
\end{equation*}
subject to $A$ being a matching (i.e., each node has at most one edge) that is maximal (i.e., no additional edge can be added).
There are known 
algorithms to solve the above problem in polynomial time (e.g., \cite{galil1986efficient}).

Note that alignments generated with the match method are inherently bidirectional. 
None of our methods require additional symmetrization as post-processing.

\subsection{Distortion and Null Extensions}

\textbf{Distortion Correction [Dist].}
Distortion, 
as introduced in IBM Model 2, is
essential for alignments based on
non-contextualized embeddings since  the similarity of two words is solely based on their surface form, independent of  position.
To penalize high distortions,
we multiply the 
similarity matrix $S$ componentwise with
\begin{align*}
P_{i,j} = 1 - \kappa\left(i/l_e-j/l_f\right)^2, 
\end{align*}
where $\kappa$ is a hyperparameter to scale the distortion matrix $P$
between $[(1 - \kappa), 1]$.
We use $\kappa = 0.5$. See 
supplementary for different values.
We can interpret this
as imposing 
a locality-preserving prior:
given a choice, a word should be aligned to a word with a
similar relative position
($\left(i/l_e-j/l_f\right)^2$
close to 0)
rather than a more distant word (large $\left(i/l_e-j/l_f\right)^2$).

\textbf{Null.}
Null words
model untranslated words and
are an important part of 
alignment models. We propose to model null words 
as follows: 
if a word is not particularly similar to any of the words in the target sentence, 
we do not align it. 
Specifically, given an alignment matrix $A$, 
we remove alignment edges when the normalized entropy of the
similarity distribution is above a threshold $\tau$, a hyperparameter.
We use normalized entropy (i.e., entropy divided by the log of sentence length) 
to account for different sentence lengths; i.e.,
we set $A_{ij} = 0$ if
$$\min ( \!- \frac{
	\sum_{k=1}^{l_f}\!\! S^h_{ik}\! \log
	S^h_{ik}}
{  \log l_f}\!,
\!- \frac{
	\sum_{k=1}^{l_e}\!\! S^v_{kj}\! \log S^v_{kj}}{\log l_e} ) \!\!>\!\! \tau,
$$
where $S^h_{ik}:= S_{ik} / \sum_{m=1}^{l_f} S_{im}$, and
$S^v_{kj}:= S_{kj} / \sum_{m=1}^{l_e} S_{mj}$.
As the ideal value of $\tau$ depends on the actual similarity scores we set 
$\tau$ to a percentile of the entropy values of the similarity distribution across all aligned edges (we use the 95th percentile).
Different percentiles are in the supplementary.

\section{Experiments}

\subsection{Embedding Learning}

\textbf{Static.} We train monolingual embeddings with
fastText  \cite{bojanowski2017enriching} for each language on its Wikipedia.
We then use  VecMap \cite{artetxe2018robust} to map the embeddings into a common multilingual space. Note 
that this algorithm works without any crosslingual
supervision (e.g., multilingual dictionaries). We use the same procedure for word and subword
levels.
We use the label  \textbf{fastText} to refer to these
embeddings as well as the alignments induced by them.

\textbf{Contextualized.} We use the multilingual BERT model (mBERT).\footnote{\label{mbertfoot}\url{https://github.com/google-research/bert/blob/master/multilingual.md}}
It is pretrained on the 104 largest Wikipedia languages. This model only provides embeddings at the subword level. 
To obtain a word embedding, we simply average the vectors of its subwords.
We consider word representations from all 12 layers as well as the concatenation of all layers. Note that 
the model is not finetuned.
We denote this method  as mBERT[i] (when using embeddings from the $i$-th layer, where $0$ means using the non-contextualized 
initial embedding layer) and mBERT[conc] (for concatenation). 

In addition, we use XLM-RoBERTa base \cite{conneau2019unsupervised}, which
is pretrained on 100 languages on cleaned CommonCrawl data \cite{wenzek2019ccnet}. We denote 
alignments obtained using the embeddings from the $i$-th
layer by XLM-R[i].

\subsection{Word and Subword Alignments}

\begin{figure}[t]
	\centering
	\begin{tabular}{c}
		\includegraphics[width=0.94\linewidth]{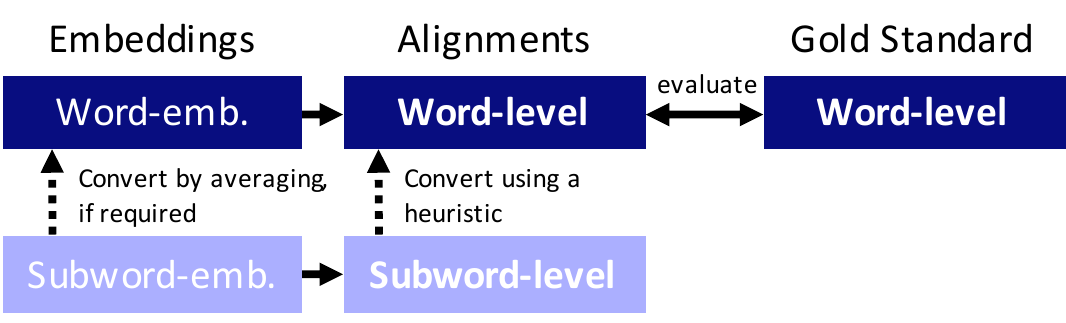}\\
		\midrule
		\includegraphics[width=0.8\linewidth]{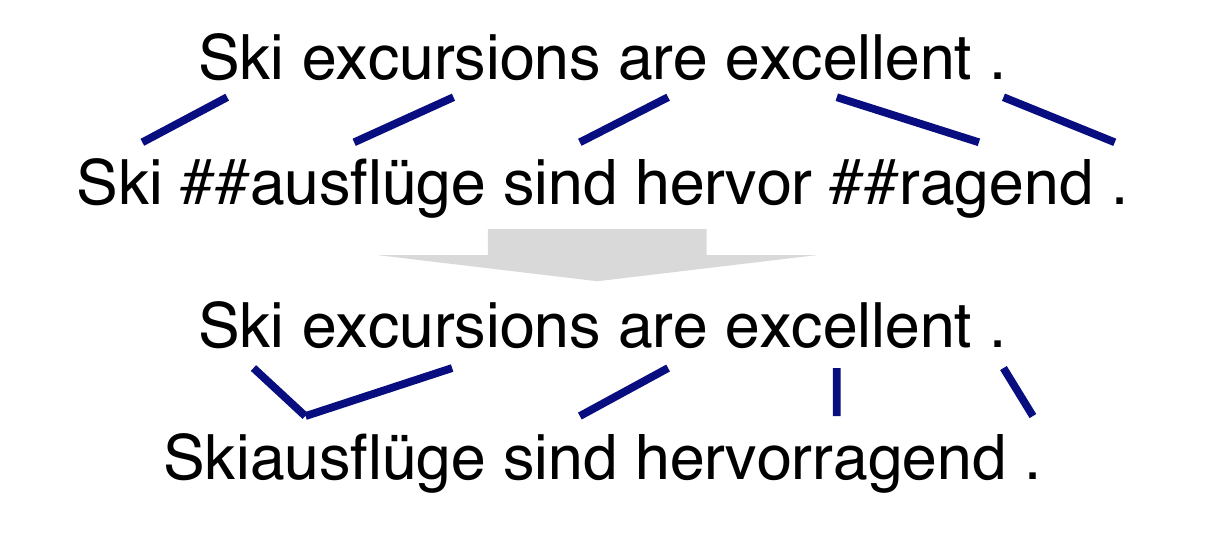}
	\end{tabular}
	\caption{Subword alignments are always converted to word alignments for evaluation.}
	\figlabel{subword2word}
\end{figure}

We investigate both alignments between
subwords such as wordpiece
\cite{schuster2012japanese}
(which are widely used for
contextualized language models) and
words. We refer to computing alignment edges between 
words as \emph{word level} and between subwords
as \emph{subword level}. Note that gold standards are all word-level. In order to 
evaluate alignments obtained at the subword level we convert subword to  word alignments 
using the heuristic ``two words are aligned if any of their  subwords are aligned'' (see \figref{subword2word}).
As a result a single word can be aligned with multiple other words.

For the
\emph{word} level, we use the NLTK tokenizer
\cite{bird2009natural} (e.g., for tokenizing Wikipedia in
order to train fastText).  For the \emph{subword} level, we
generally use multilingual BERT's vocabulary\footnotemark[3] and BERT's
wordpiece tokenizer. For XLM-R we use the XLM-R subword
vocabulary.
Since gold standards are already tokenized,
they do not require additional tokenization.

\subsection{Baselines}
We compare to three popular statistical alignment models that all require parallel training data.
\textbf{fast-align/IBM2} \cite{dyer2013simple} is an implementation of an alignment algorithm based on
IBM Model 2. It is popular
because of
its speed and high quality. \textbf{eflomal}\footnote{\url{github.com/robertostling/eflomal}}
(based on efmaral by \newcite{ostling2016efficient}), a 
Bayesian model with Markov Chain Monte Carlo inference,  is claimed to outperform fast-align on speed and quality.
Further we use the widely used software package \textbf{Giza++/IBM4} \cite{och03:asc}, which implements IBM alignment models.
We use its standard settings: 5 iterations each for the HMM model, IBM Models 1, 3 and 4 with $p_0=0.98$.

\textbf{Symmetrization.}
Probabilistic word alignment models create forward and
backward alignments and then symmetrize them \cite{och03:asc,koehn2005edinburgh}.
We compared the symmetrization methods 
grow-diag-final-and (GDFA) and intersection and found them to 
perform comparably; see supplementary.
We use GDFA throughout the paper.

\subsection{Evaluation Measures}
Given a set of predicted alignment edges $A$ and a set of sure, possible gold standard edges $S$, $P$ (where $S\subset P$),
we use the following evaluation measures: 
\begin{align*}
\text{prec} &= \frac{|A \cap P|}{|A|},
\text{rec} = \frac{|A \cap S|}{|S|},\\
F_1 &= \frac{2 \; \text{prec}\; \text{rec}}{\text{prec} + \text{rec}},\\
\text{AER}  &= 1 - \frac{|A \cap S| + |A \cap P|}{|A| + |S|},
\end{align*}
where $|\cdot|$ denotes the cardinality of a set.
This is the standard  evaluation \cite{och03:asc}.

\subsection{Data}

\begin{table*}[t]
	\centering
	\begin{threeparttable}
		\centering
		\scriptsize
		\begin{tabular}{l||rrrr|rr|r}
			& Gold & Gold St. & && Parallel & Parallel  & Wikipedia \\
			Lang. & Standard & Size & $|S|$&$|P\setminus S|$&Data & Data Size & Size \\
			\midrule
			ENG-CES & \cite{marecek:2008} & 2500 &44292&23132&EuroParl \cite{koehn2005europarl} & 646k & 8M \\
			ENG-DEU & EuroParl-based\tnote{a}   &  508 &9612&921&  EuroParl \cite{koehn2005europarl} & 1920k & 48M \\
			ENG-FAS & \cite{tavakoli2014phrase} &  400 &11606&0& TEP \cite{pilevar2011tep} & 600k & 5M\\
			ENG-FRA & WPT2003, \cite{och2000improved}, & 447 &4038&13400& Hansards \cite{germann2001aligned} & 1130k & 32M\\
			ENG-HIN & WPT2005\tnote{b} & 90 &1409&0& Emille \cite{mcenery2000emille} & 3k & 1M\\
			ENG-RON & WPT2005\tnote{b}& 203 &5033&0& Constitution, Newspaper\tnote{b} & 50k & 3M\\
		\end{tabular}
		\begin{tablenotes}
			\item[a]  \url{www-i6.informatik.rwth-aachen.de/goldAlignment/} 
			\item[b] \url{http://web.eecs.umich.edu/~mihalcea/wpt05/}
		\end{tablenotes}
	\end{threeparttable}
	\caption{Overview of datasets. ``Lang.'' uses ISO 639-3 language codes. ``Size'' refers to
		the number of sentences. ``Parallel Data Size'' refers to the number 
		of parallel sentences in addition to the gold alignments that is used for training the baselines. Our sentence tokenized version of 
		the English Wikipedia has 105M sentences. \tablabel{data}}
\end{table*}

Our \textbf{test data} are a diverse set of 6 language pairs: 
Czech, German, Persian, French, Hindi and Romanian, always paired with English.
See \tabref{data} for  corpora and supplementary for URLs.

For our baselines requiring parallel training data (i.e., eflomal, fast-align and Giza++) 
we select additional parallel \textbf{training data} that is consistent
with the target domain where available. See \tabref{data} for 
the corpora. Unless indicated otherwise
we use the whole parallel training data.
\figref{learningcurve}
shows the effect of using more or less training data.

Given the large amount of possible experiments when considering
6 language pairs we do not have space to present all numbers for all languages.
If we show results for only one pair, we choose ENG-DEU as it is an established and well-known
dataset (EuroParl). If we show results for more languages we fall back to DEU, CES and HIN, to show
effects on a mid-resource morphologically rich language (CES) and a low-resource language 
written in a different script (HIN).

\section{Results}

\subsection{Embedding Layer}

\begin{figure}[t]
	\centering
	\includegraphics[width=1.0\linewidth]{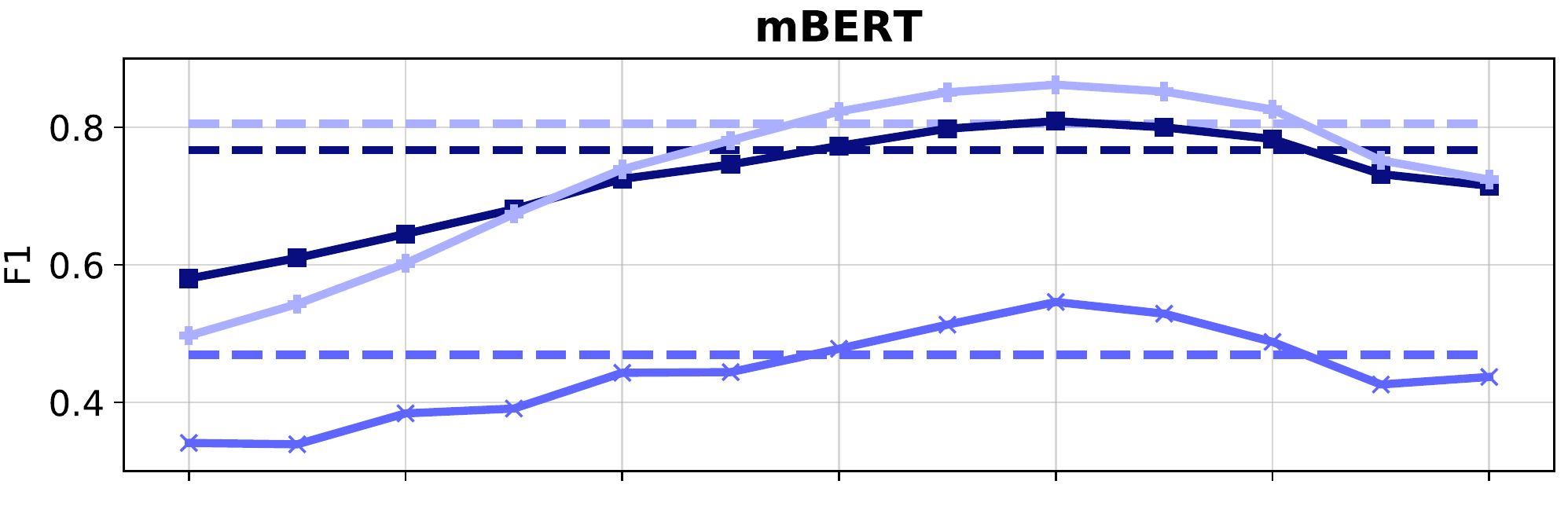}
	\includegraphics[width=1.0\linewidth]{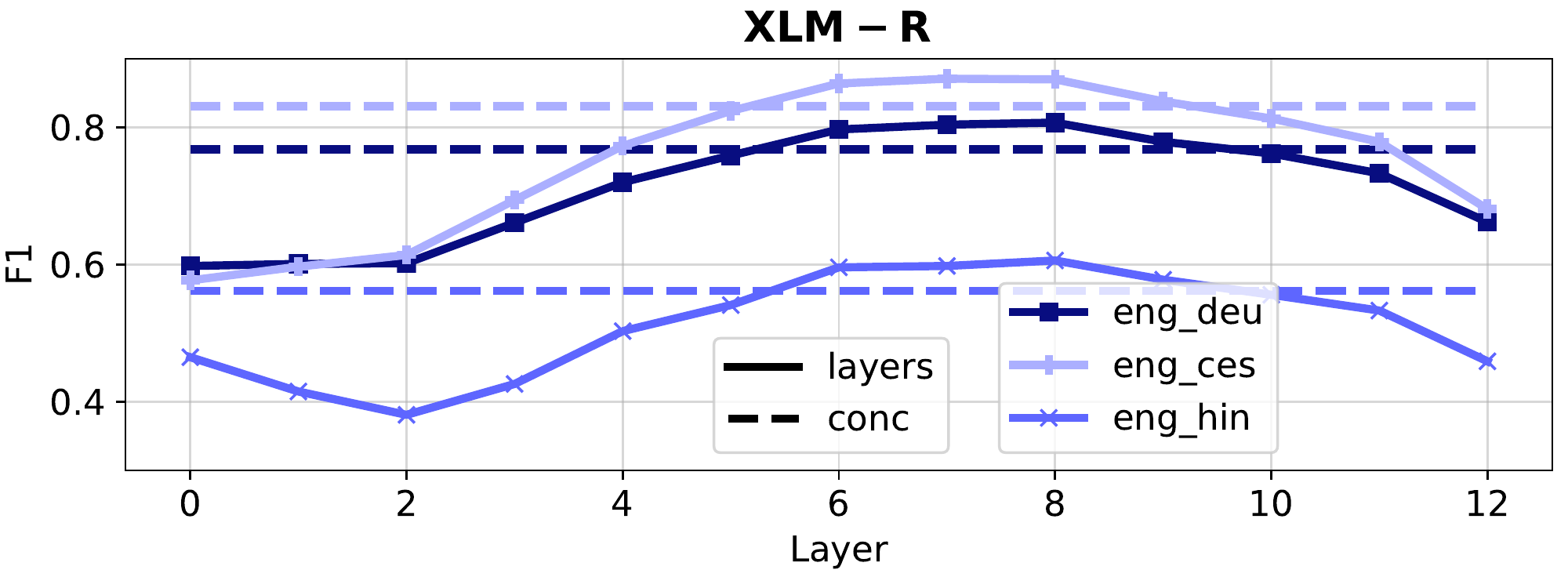}
	\caption{Word alignment performance across layers of mBERT (top) and XLM-R (bottom). 
		Results are $F_1$ with Argmax at the subword
                level.}
                \figlabel{bertlayer}
\end{figure}

\figref{bertlayer} shows a parabolic
trend across layers of mBERT and XLM-R. We use layer 8
in this paper because it has
best
performance. This is consistent with other work \cite{hewitt-manning-2019-structural,tenney2019bert}: 
in the first layers the contextualization is too weak for 
high-quality alignments while the last layers are too specialized on the pretraining task (masked language modeling).

\subsection{Comparison with Prior Work}

\begin{table*}[t]
	\scriptsize
	\centering
	\def\tablesep{0.02cm}
	\begin{tabular}{
			@{\hspace{\tablesep}}c@{\hspace{\tablesep}}
			@{\hspace{\tablesep}}c@{\hspace{\tablesep}}|
			@{\hspace{\tablesep}}l@{\hspace{\tablesep}}||
			@{\hspace{\tablesep}}c@{\hspace{\tablesep}}
			@{\hspace{\tablesep}}c@{\hspace{\tablesep}}|
			@{\hspace{\tablesep}}c@{\hspace{\tablesep}}
			@{\hspace{\tablesep}}c@{\hspace{\tablesep}}|
			@{\hspace{\tablesep}}c@{\hspace{\tablesep}}
			@{\hspace{\tablesep}}c@{\hspace{\tablesep}}|
			@{\hspace{\tablesep}}c@{\hspace{\tablesep}}
			@{\hspace{\tablesep}}c@{\hspace{\tablesep}}|
			@{\hspace{\tablesep}}c@{\hspace{\tablesep}}
			@{\hspace{\tablesep}}c@{\hspace{\tablesep}}|
			@{\hspace{\tablesep}}c@{\hspace{\tablesep}}
			@{\hspace{\tablesep}}c@{\hspace{\tablesep}}}
		&& & \multicolumn{2}{c}{ ENG-CES } & \multicolumn{2}{c}{ ENG-DEU } & \multicolumn{2}{c}{ ENG-FAS } & \multicolumn{2}{c}{ ENG-FRA } & \multicolumn{2}{c}{ ENG-HIN } & \multicolumn{2}{c}{ ENG-RON } \\
		&& Method & $F_1$ & AER & $F_1$ & AER & $F_1$ & AER & $F_1$ & AER & $F_1$ & AER & $F_1$ & AER \\
		\midrule
\multirow{7}{*}{ \rotatebox{90}{Prior Work} }
& & \cite{ostling2015bayesian} Bayesian & & & & & & & \textbf{.94 }& \textbf{.06 }& .57 & .43 & \textbf{.73 }& \textbf{.27 }\\
& & \cite{ostling2015bayesian} Giza++ & & & & & & & .92 & .07 & .51 & .49 & .72 & .28 \\
& & \cite{legrand2016neural} Ensemble Method & .81 & .16 & & & & & .71 & .10 & & & &  \\
& & \cite{ostling2016efficient} efmaral & & & & & & & .93 & .08 & .53 & .47 & .72 & .28 \\
& & \cite{ostling2016efficient} fast-align & & & & & & & .86 & .15 & .33 & .67 & .68 & .33 \\
& & \cite{zenkel2019adding} Giza++ & & & & .21 & & & & \textbf{.06} & & & & .28 \\
& & \cite{garg2019jointly} Multitask & & & & .20 & & & & .08 & & & &  \\
\midrule
\multirow{6}{*}{ \rotatebox{90}{Baselines} } & \multirow{3}{*}{ \rotatebox{90}{Word} } & fast-align/IBM2 & .76 & .25 & .71 & .29 & .57 & .43 & .86 & .15 & .34 & .66 & .68 & .33 \\
& & Giza++/IBM4 & .75 & .26 & .77 & .23 & .51 & .49 & .92 & .09 & .45 & .55 & .69 & .31 \\
& & eflomal & .85 & .15 & .77 & .23 & .61 & .39 & .93 & .08 & .51 & .49 & .71 & .29 \\
\cmidrule{2-15}
& \multirow{3}{*}{ \rotatebox{90}{Subword} } & fast-align/IBM2 & .78 & .23 & .71 & .30 & .58 & .42 & .85 & .16 & .38 & .62 & .68 & .32 \\
& & Giza++/IBM4 & .82 & .18 & .78 & .22 & .57 & .43 & .92 & .09 & .48 & .52 & .69 & .32 \\
& & eflomal & .84 & .17 & .76 & .24 & .63 & .37 & .91 & .09 & .52 & .48 & .72 & .28 \\
\midrule
\multirow{6}{*}{ \rotatebox{90}{This Work} }
& \multirow{3}{*}{ \rotatebox{90}{Word} }
& fastText - Argmax & .70 & .30 & .60 & .40 & .50 & .50 & .77 & .22 & .49 & .52 & .47 & .53 \\
& & mBERT[8] - Argmax & \textbf{.87 }& \textbf{.13 }& .79 & .21 & .67 & .33 & \textbf{.94} & \textbf{.06 }& .54 & .47 & .64 & .36 \\
& & XLM-R[8] - Argmax & \textbf{.87 }& \textbf{.13 }& .79 & .21 & .70 & .30 & .93 & \textbf{.06 }& .59 & .41 & .70 & .30 \\
\cmidrule{2-15}
& \multirow{3}{*}{ \rotatebox{90}{Subword} }
& fastText - Argmax & .58 & .42 & .56 & .44 & .09 & .91 & .73 & .26 & .04 & .96 & .43 & .58 \\
& & mBERT[8] - Argmax & .86 & .14 & \textbf{.81 }& \textbf{.19 }& .67 & .33 & \textbf{.94 }& \textbf{.06 }& .55 & .45 & .65 & .35 \\
& & XLM-R[8] - Argmax & \textbf{.87 }& \textbf{.13 }& \textbf{.81 }& \textbf{.19 }& \textbf{.71 }& \textbf{.29 }& .93 & .07 & \textbf{.61 }& \textbf{.39 }& .71 & .29 \\
	\end{tabular}
	\caption{Comparison of our methods, baselines and prior work in unsupervised word alignment. Best result per column in bold. A detailed version of the table with precision/recall and Itermax/Match results is in supplementary. }
	\tablabel{main}
\end{table*}

\textbf{Contextual Embeddings.} 
\tabref{main} shows that mBERT and XLM-R consistently perform well with the Argmax method. 
XLM-R
yields mostly higher values than mBERT. 
Our three baselines, eflomal, fast-align and Giza++, are always outperformed (except for RON). 
We outperform all prior work except for FRA where we match
the performance and RON.
This comparison is not entirely fair because
methods 
relying on parallel data have access to
the parallel sentences of the test data
during training whereas our methods do not.

Romanian might be a special 
case as it exhibits a large amount of many to one links and further lacks determiners. 
How determiners are handled in the gold standard depends heavily on the annotation guidelines. 
Note that one of our settings, XLM-R[8] with Itermax
at the subword level, has
an F1 of .72 for ENG-RON, 
which comes very close to the performance 
by \cite{ostling2015bayesian} (see \tabref{methodoverview}).

In summary, extracting alignments from similarity matrices
is a very simple and efficient method that performs
surprisingly strongly. It outperforms strong statistical baselines and most prior work in unsupervised word alignment 
for
CES, DEU, FAS and HIN and is comparable for FRA and RON.
We attribute this 
to the strong contextualization in mBERT and XLM-R.

\textbf{Static Embeddings.} 
fastText shows a solid performance on word level, which is worse but comes close to fast-align and outperforms 
it for HIN. We consider this surprising as fastText did not have access
to parallel data or any multilingual signal. VecMap can also be used with 
crosslingual dictionaries. We expect this to boost
performance and
fastText could then become a viable alternative to fast-align.

\begin{figure}[t]
	\centering
	\includegraphics[width=1.0\linewidth]{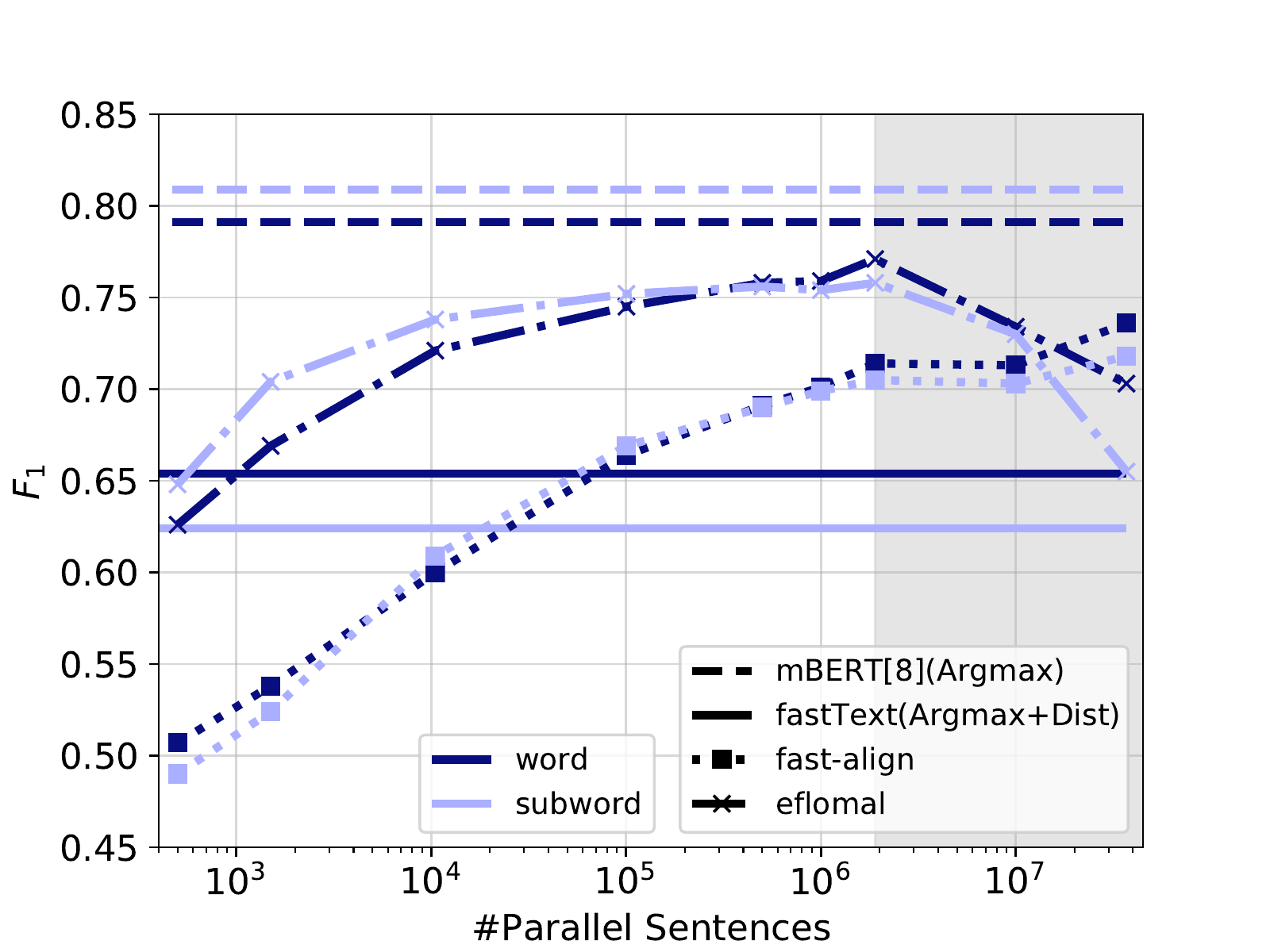}
	\caption{Learning curves of fast-align/eflomal
		vs.\ embedding-based alignments.
		Results shown are $F_1$ for ENG-DEU, contrasting subword and word representations.
		Up to 1.9M parallel sentences we use EuroParl. To demonstrate the effect
		with abundant parallel data we add up to 37M \emph{additional} parallel sentences from ParaCrawl \cite{espla2019paracrawl}
		(see grey area).}
	\figlabel{learningcurve}
\end{figure}

\textbf{Amount of Parallel Data.}  \figref{learningcurve}
shows that fast-align and eflomal get better with more
training data with eflomal outperforming fast-align, as
expected. However, even with 1.9M parallel sentences mBERT
outperforms both baselines. When adding up to 37M additional
parallel sentences from ParaCrawl \cite{espla2019paracrawl}
performance for fast-align increases slightly, however,
eflomal decreases (grey area in plot).  ParaCrawl contains mined parallel
sentences whose lower quality 
probably  harms eflomal.  fastText (with distortion) is
competitive with eflomal for fewer than 1000 parallel
sentences and outperforms fast-align even with 10k
sentences. Thus for very small parallel corpora ($<$10k
sentences) using fastText embeddings is an alternative to
fast-align.

\emph{The main takeaway from \figref{learningcurve} is that
	mBERT-based alignments, a method that does not
	need any parallel training data, outperforms state-of-the-art aligners
	like eflomal
	for ENG-DEU, even in the very high resource case.}

\subsection{Additional Methods and Extensions}
\seclabel{hyperparams}

\begin{table}[t]
	\def\symmsep{0.1cm}
	\scriptsize
	\centering
	\begin{tabular}{@{\hspace{\symmsep}}l@{\hspace{\symmsep}}
			@{\hspace{\symmsep}}l@{\hspace{\symmsep}}||
			@{\hspace{\symmsep}}c@{\hspace{\symmsep}}
			@{\hspace{\symmsep}}c@{\hspace{\symmsep}}
			@{\hspace{\symmsep}}c@{\hspace{\symmsep}}
			@{\hspace{\symmsep}}c@{\hspace{\symmsep}}
			@{\hspace{\symmsep}}c@{\hspace{\symmsep}}
			@{\hspace{\symmsep}}c@{\hspace{\symmsep}}}
		& & ENG- & ENG- & ENG- & ENG- & ENG- & ENG- \\
		Emb. & Method & CES & DEU & FAS & FRA & HIN & RON \\
		\midrule
		\midrule
		\multirow{3}{*}{ mBERT[8] } & Argmax & \textbf{.86} & \textbf{.81} & .67 & \textbf{.94} & .55 & .65 \\
		& Itermax & \textbf{.86} & \textbf{.81} & \textbf{.70}& .93 & \textbf{.58} & \textbf{.69} \\
		& Match & .82 & .78 & .67 & .90& \textbf{.58} & .67 \\
		\midrule
		\multirow{3}{*}{ XLM-R[8] } & Argmax & \textbf{.87} & \textbf{.81} & .71 & \textbf{.93} & .61& .71 \\
		& Itermax & .86 & .80& \textbf{.72} & .92 & \textbf{.62} & \textbf{.72} \\
		& Match & .81 & .76 & .68 & .88 & .60& .70\\
	\end{tabular}
	\caption{Comparison of our three proposed methods across all languages for the best embeddings from \tabref{main}: mBERT[8] and XLM-R[8]. We show $F_1$ at the subword level.
		Best result per embedding type in bold. \tablabel{methodoverview}}
\end{table}

\begin{table}[t]
	\scriptsize
	\def\symmsep{0.04cm}
	\begin{tabular}{@{\hspace{\symmsep}}l@{\hspace{\symmsep}}@{\hspace{\symmsep}}c@{\hspace{\symmsep}}@{\hspace{\symmsep}}c@{\hspace{\symmsep}}||@{\hspace{\symmsep}}c@{\hspace{\symmsep}}@{\hspace{\symmsep}}c@{\hspace{\symmsep}}@{\hspace{\symmsep}}c@{\hspace{\symmsep}}@{\hspace{\symmsep}}c@{\hspace{\symmsep}}|@{\hspace{\symmsep}}c@{\hspace{\symmsep}}@{\hspace{\symmsep}}c@{\hspace{\symmsep}}@{\hspace{\symmsep}}c@{\hspace{\symmsep}}@{\hspace{\symmsep}}c@{\hspace{\symmsep}}|@{\hspace{\symmsep}}c@{\hspace{\symmsep}}@{\hspace{\symmsep}}c@{\hspace{\symmsep}}@{\hspace{\symmsep}}c@{\hspace{\symmsep}}@{\hspace{\symmsep}}c@{\hspace{\symmsep}}}
		& & & \multicolumn{4}{c}{ ENG-DEU } & \multicolumn{4}{c}{ ENG-CES } & \multicolumn{4}{c}{ ENG-HIN } \\
		\rotatebox{90}{Emb.} & $n_{\text{max}}$ & $\alpha$ & Prec. & Rec. & $F_1$ & AER & Prec. & Rec. & $F_1$ & AER & Prec. & Rec. & $F_1$ & AER \\
		\midrule
		\multirow{7}{*}{\rotatebox{90}{mBERT[8]} } & 1 & - & \textbf{.92 }& .69 & .79 & .21 & \textbf{.95 }& .80 & \textbf{.87 }& \textbf{.13 }& \textbf{.84 }& .39 & .54 & .47 \\
		\cmidrule{2-15}
		& \multirow{3}{*}{ 2 } & .90& .85 & .77 & \textbf{.81 }& \textbf{.19 }& .87 & .87 & \textbf{.87 }& .14 & .75 & .47 & .58 & .42 \\
		& & .95 & .83 & .80 & \textbf{.81 }& \textbf{.19 }& .85 & .89 & \textbf{.87 }& \textbf{.13 }& .73 & .48 & .58 & .42 \\
		& & 1 & .77 & .79 & .78 & .22 & .80 & .86 & .83 & .17 & .63 & .46 & .53 & .47 \\
		\cmidrule{2-15}
		& \multirow{3}{*}{ 3 } & .90& .81 & .80 & .80 & .20 & .83 & .88 & .85 & .15 & .70 & .49 & .57 & .43 \\
		& & .95 & .78 & \textbf{.83 }& \textbf{.81 }& .20 & .81 & \textbf{.91 }& .86 & .15 & .68 & \textbf{.52 }& \textbf{.59 }& \textbf{.41 }\\
		& & 1 & .73 & \textbf{.83 }& .77 & .23 & .76 & \textbf{.91 }& .82 & .18 & .58 & .51 & .54 & .46 \\
		\midrule
		\multirow{7}{*}{ \rotatebox{90}{fastText} } & 1& - & \textbf{.81 }& .48 & .60 & .40 & \textbf{.86 }& .59 & .70 & .30 & \textbf{.75 }& .36 & .49 & .52 \\
		\cmidrule{2-15}
		& \multirow{3}{*}{ 2 } & .90& .69 & .56 & \textbf{.62 }& \textbf{.38 }& .74 & .69 & \textbf{.72 }& \textbf{.29 }& .63 & .42 & \textbf{.51 }& \textbf{.49 }\\
		& & .95 & .66 & .56 & .61 & .39 & .71 & .69 & .70 & .30 & .59 & .41 & .48 & .52 \\
		& & 1 & .59 & .55 & .57 & .43 & .62 & .65 & .63 & .37 & .53 & .39 & .45 & .55 \\
		\cmidrule{2-15}
		& \multirow{3}{*}{ 3 } & .90& .63 & \textbf{.59 }& .61 & .39 & .67 & .72 & .70 & .31 & .57 & .43 & .49 & .51 \\
		& & .95 & .59 & \textbf{.59 }& .59 & .41 & .63 & \textbf{.73 }& .68 & .33 & .53 & \textbf{.44 }& .48 & .52 \\
		& & 1 & .53 & .58 & .55 & .45 & .55 & .70 & .62 & .39 & .48 & .43 & .45 & .55 \\
	\end{tabular}
	\caption{Itermax with different number of iterations ($n_{\text{max}}$) and different $\alpha$. Results are at the word level. \tablabel{itermaxresult}}
\end{table}

We already showed that Argmax yields alignments that are competitive with the state of the art. 
In this section we compare all our proposed methods and extensions more closely.

\begin{table}[t]
	\scriptsize
	\centering
	\def\postprocsep{0.04cm}
	\begin{tabular}{@{\hspace{\postprocsep}}c@{\hspace{\postprocsep}}|@{\hspace{\postprocsep}}l@{\hspace{\postprocsep}}||@{\hspace{\postprocsep}}c@{\hspace{\postprocsep}}@{\hspace{\postprocsep}}c@{\hspace{\postprocsep}}@{\hspace{\postprocsep}}c@{\hspace{\postprocsep}}@{\hspace{\postprocsep}}c@{\hspace{\postprocsep}}|@{\hspace{\postprocsep}}c@{\hspace{\postprocsep}}@{\hspace{\postprocsep}}c@{\hspace{\postprocsep}}@{\hspace{\postprocsep}}c@{\hspace{\postprocsep}}@{\hspace{\postprocsep}}c@{\hspace{\postprocsep}}|@{\hspace{\postprocsep}}c@{\hspace{\postprocsep}}@{\hspace{\postprocsep}}c@{\hspace{\postprocsep}}@{\hspace{\postprocsep}}c@{\hspace{\postprocsep}}@{\hspace{\postprocsep}}c@{\hspace{\postprocsep}}}
		& & \multicolumn{4}{c}{ ENG-DEU } & \multicolumn{4}{c}{ ENG-CES } & \multicolumn{4}{c}{ ENG-HIN } \\
		\rotatebox{90}{Emb.} & Method & Prec. & Rec. & $F_1$ & AER & Prec. & Rec. & $F_1$ & AER & Prec. & Rec. & $F_1$ & AER \\
		\midrule
		\midrule
		\multirow{9}{*}{ \rotatebox{90}{fastText} } & Argmax & .81 & .48 & .60 & .40 & .86 & .59 & .70 & .30 & \textbf{.75 }& \textbf{.36 }& \textbf{.49 }& \textbf{.52 }\\
		& \emph{+Dist} & \textbf{.84 }& \textbf{.54 }& \textbf{.65 }& \textbf{.35 }& \textbf{.89 }& \textbf{.68 }& \textbf{.77 }& \textbf{.23 }& .64 & .30 & .41 & .59 \\
		& \emph{+Null} & .81 & .46 & .59 & .41 & .86 & .56 & .68 & .32 & .74 & .34 & .46 & .54 \\
		\cmidrule{2-14}
		& Itermax & .69 & .56 & .62 & .38 & .74 & .69 & .72 & .29 & \textbf{.63} & \textbf{.42 }& \textbf{.51 }& \textbf{.49 }\\
		& \emph{+Dist} & \textbf{.71 }& \textbf{.62 }& \textbf{.66 }& \textbf{.34 }& \textbf{.75 }& \textbf{.76 }& \textbf{.76 }& \textbf{.25 }& .54 & .37 & .44 & .57 \\
		& \emph{+Null} & .69 & .53 & .60 & .40 & .74 & .66 & .70 & .30 & \textbf{.63 }& .40 & .49 & .51 \\
		\cmidrule{2-14}
		& Match & .60 & .58 & .59 & .41 & .65 & .71 & .68 & .32 & .55 & \textbf{.43 }& \textbf{.48 }& \textbf{.52 }\\
		& \emph{+Dist} & \textbf{.67 }& \textbf{.64 }& \textbf{.65 }& \textbf{.35 }& \textbf{.72 }& \textbf{.78 }& \textbf{.75 }& \textbf{.25 }& .50 & .39 & .43 & .57 \\
		& \emph{+Null} & .61 & .56 & .58 & .42 & .66 & .69 & .67 & .33 & \textbf{.56 }& .41 & \textbf{.48} & \textbf{.52} \\
		\midrule
		\multirow{9}{*}{ \rotatebox{90}{mBERT[8]} } & Argmax & .92 & \textbf{.69 }& \textbf{.79 }& \textbf{.21 }& \textbf{.95 }& \textbf{.80 }& \textbf{.87 }& \textbf{.13 }& .84 & \textbf{.39 }& \textbf{.54 }& \textbf{.47 }\\
		& \emph{+Dist} & .91 & .67 & .77 & .23 & .93 & .79 & .85 & .15 & .68 & .29 & .41 & .59 \\
		& \emph{+Null} & \textbf{.93 }& .67 & .78 & .22 & \textbf{.95 }& .77 & .85 & .15 & \textbf{.85 }& .38 & .53 & \textbf{.47} \\
		\cmidrule{2-14}
		& Itermax & .85 & \textbf{.77 }& \textbf{.81 }& \textbf{.19 }& .87 & \textbf{.87 }& \textbf{.87 }& \textbf{.14 }& .75 & \textbf{.47 }& \textbf{.58 }& \textbf{.43 }\\
		& \emph{+Dist} & .82 & .75 & .79 & .21 & .84 & .85 & .85 & .15 & .56 & .34 & .43 & .58 \\
		& \emph{+Null} & \textbf{.86 }& .75 & .80 & .20 & \textbf{.88 }& .84 & .86 & \textbf{.14} & \textbf{.76 }& .45 & .57 & \textbf{.43} \\
		\cmidrule{2-14}
		& Match & .78 & \textbf{.74 }& \textbf{.76 }& \textbf{.24 }& .81 & \textbf{.85 }& \textbf{.83 }& \textbf{.17 }& .67 & \textbf{.52 }& \textbf{.59 }& \textbf{.42 }\\
		& \emph{+Dist} & .75 & .71 & .73 & .27 & .79 & .83 & .81 & .20 & .45 & .35 & .39 & .61 \\
		& \emph{+Null} & \textbf{.80 }& .73 & \textbf{.76 }& \textbf{.24 }& \textbf{.83 }& .83 & \textbf{.83 }& \textbf{.17 }& \textbf{.68 }& .51 & .58 & \textbf{.42} \\
	\end{tabular}
	\caption{Analysis of Null and Distortion Extensions. All alignments are obtained at word-level. Best result per embedding type and method  in bold.}
	\tablabel{postproc}
\end{table}

\textbf{Itermax.}
\tabref{itermaxresult} shows results for Argmax (i.e., 1 Iteration) as well as Itermax (i.e., 
2 or more iterations of Argmax). As expected, with more iterations precision drops in favor of recall. 
Overall, Itermax achieves higher $F_1$ scores for the three
language pairs (equal for ENG-CES) both for mBERT[8] and fastText embeddings.
For Hindi the performance increase is the highest. We hypothesize that for more distant languages
Itermax is more beneficial as similarity between wordpieces may be generally lower, 
thus exhibiting fewer mutual argmaxes. For the rest of the paper if we use Itermax we use
2 Iterations with $\alpha=0.9$ as it exhibits best performance (5 out of 6 wins in \tabref{itermaxresult}).

\textbf{Argmax/Itermax/Match.}
In \tabref{methodoverview} we compare our three proposed methods in terms of $F_1$ across all languages. 
We chose to show the two best performing settings from \tabref{main}: mBERT[8] and XLM-R[8] at the subword level. 
Itermax performs slightly better than Argmax with 6 wins, 4 losses and 2 ties. Itermax seems to help more for 
more distant languages such as FAS, HIN and RON, but harms for FRA. Match has the lowest $F_1$, but generally exhibits a higher recall (see e.g., \tabref{postproc}).

\textbf{Null and Distortion Extensions.}
\tabref{postproc} shows that 
Argmax and Itermax generally have higher precision, whereas Match has higher recall. 
Adding Null almost always increases precision, but at the cost of recall, resulting mostly in a 
lower $F_1$ score. Adding a distortion prior boosts performance
for static embeddings, e.g., from .70 to .77 for ENG-CES
Argmax $F_1$
and similarly for ENG-DEU. 
For Hindi a distortion prior is harmful.  
Dist has little and sometimes harmful effects on mBERT indicating that mBERT's
contextualized representations already
match well across languages.

\textbf{Summary.} Argmax and Itermax exhibit the best and most stable performance.
For most language pairs Itermax is recommended.
If high recall alignments are required,  Match is the recommended algorithm.
Except for HIN, a distortion prior is beneficial
for static embeddings. Null should be applied when
one wants to push precision even higher (e.g., for annotation projection).

\subsection{Words and Subwords}

\begin{figure}[t]
	\centering
	\includegraphics[width=1.0\linewidth]{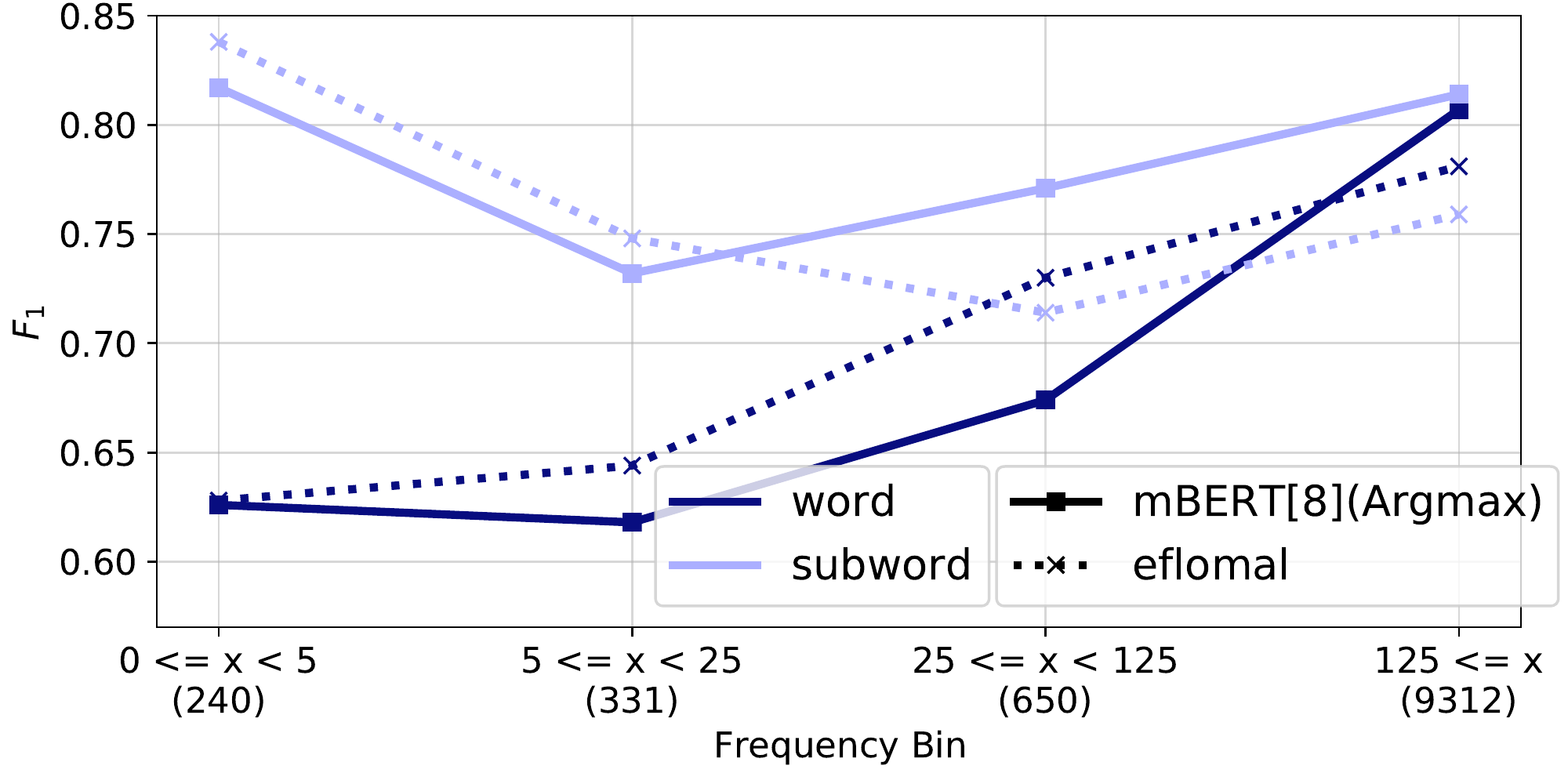}
	\caption{ Results for different frequency bins on ENG-DEU. An edge in $S$, $P$, or $A$ 
		is attributed to exactly one bin based on the minimum frequency of the involved words (denoted by $x$). Number of gold
		edges in brackets.
		Eflomal is trained on all 1.9M parallel sentences. Frequencies are computed on the same corpus. 
	}
	\figlabel{rarewords}
\end{figure}

\tabref{main} shows that subword processing  slightly outperforms
word-level processing for most methods. Only 
fastText is harmed by subword processing. 
We use VecMap to match
(sub)word distributions across languages. 
We hypothesize that it is harder to match subword than word distributions --
this effect is strongest for Persian and Hindi, probably due to different scripts and thus  different subword distributions. 
Initial experiments showed that adding supervision in form of a dictionary helps restore performance. 
We will investigate this in future work.

We hypothesize that subword processing is beneficial for
aligning rare words. To show this, we compute our evaluation measures for
different frequency bins. More specifically,
we
only consider gold standard alignment edges for the computation where at least one of the
member words has a certain frequency in a reference corpus (in our case all 1.9M lines
from the ENG-DEU EuroParl corpus). 
That is, we only consider
the edge $(i,j)$ in $A,S$ or $P$ if the minimum of the source and 
target word frequency is
in $[\gamma_{l},\gamma_{u})$ where 
$\gamma_{l}$ and $\gamma_{u}$ are bin boundaries.

\figref{rarewords} shows $F_1$ for different frequency bins. For rare words
both eflomal and mBERT show a severely decreased performance at the word level, but not 
at the subword level. Thus, subword processing is indeed
beneficial for rare words.

\subsection{Part-Of-Speech Analysis}

\begin{table}
	\def\symmsep{0.08cm}
	\scriptsize
	\centering
	\begin{tabular}{@{\hspace{\symmsep}}l@{\hspace{\symmsep}}
		@{\hspace{\symmsep}}l@{\hspace{\symmsep}}||
		@{\hspace{\symmsep}}c@{\hspace{\symmsep}}
		@{\hspace{\symmsep}}c@{\hspace{\symmsep}}
		@{\hspace{\symmsep}}c@{\hspace{\symmsep}}
		@{\hspace{\symmsep}}c@{\hspace{\symmsep}}
		@{\hspace{\symmsep}}c@{\hspace{\symmsep}}
		@{\hspace{\symmsep}}c@{\hspace{\symmsep}}
		@{\hspace{\symmsep}}c@{\hspace{\symmsep}}}
	 & & ADJ & ADP & ADV & AUX & NOUN & PRON & VERB \\
	\midrule
	\midrule
	\multirow{2}{*}{eflomal} & Word & \textbf{0.83} & 0.69 & \textbf{0.72} & 0.63 & 0.85 & 0.79 & 0.63 \\
	& Subword & 0.82 & 0.68 & 0.71 & 0.57 & 0.85 & 0.77 & 0.62 \\
	\midrule
	\multirow{2}{*}{mBERT[8]} & Word & 0.79 & 0.74 & 0.71 & 0.71 & 0.81 & \textbf{0.84} & \textbf{0.69} \\
	& Subword & 0.81 & \textbf{0.75} & \textbf{0.72} & \textbf{0.72} & \textbf{0.87} & \textbf{0.84} & \textbf{0.69} 

\end{tabular}
        \caption{Alignment performance ($F_1$) on ENG-DEU for POS.
We use mBERT[8](Argmax) and
          Eflomal trained on 1.9M parallel sentences on the
          word level.\tablabel{pos_table}}
\end{table}

\begin{figure}[t]
	\centering
	\includegraphics[width=1.0\linewidth]{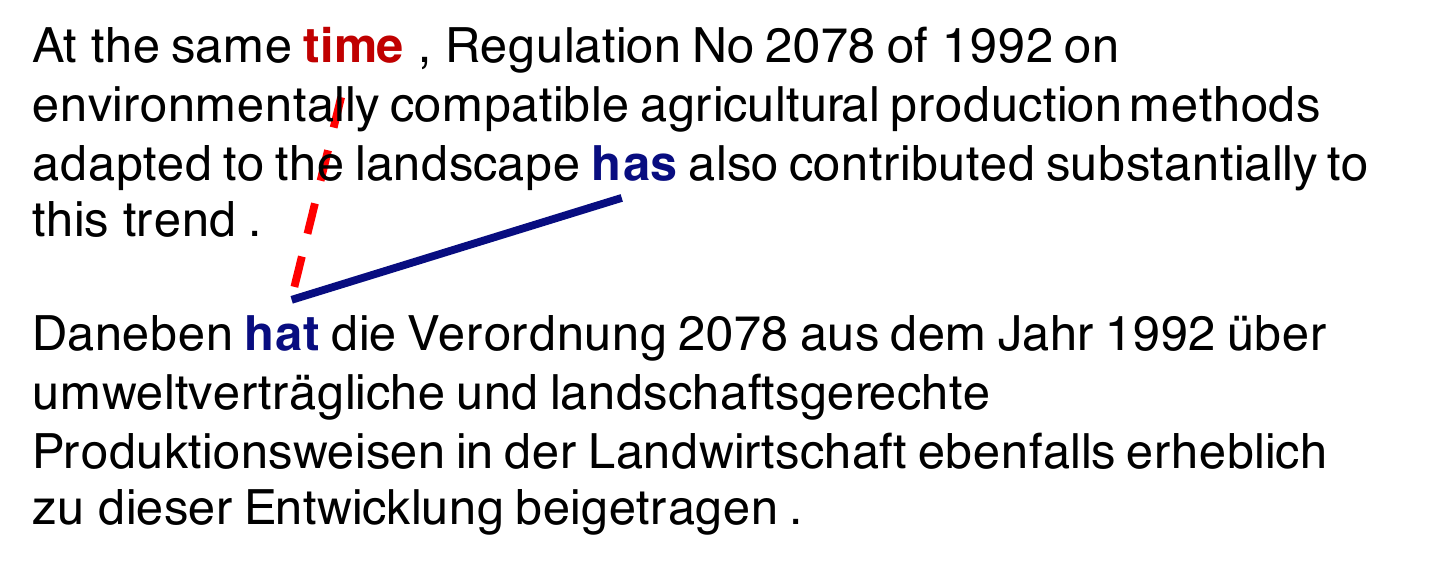}
	\includegraphics[width=1.0\linewidth]{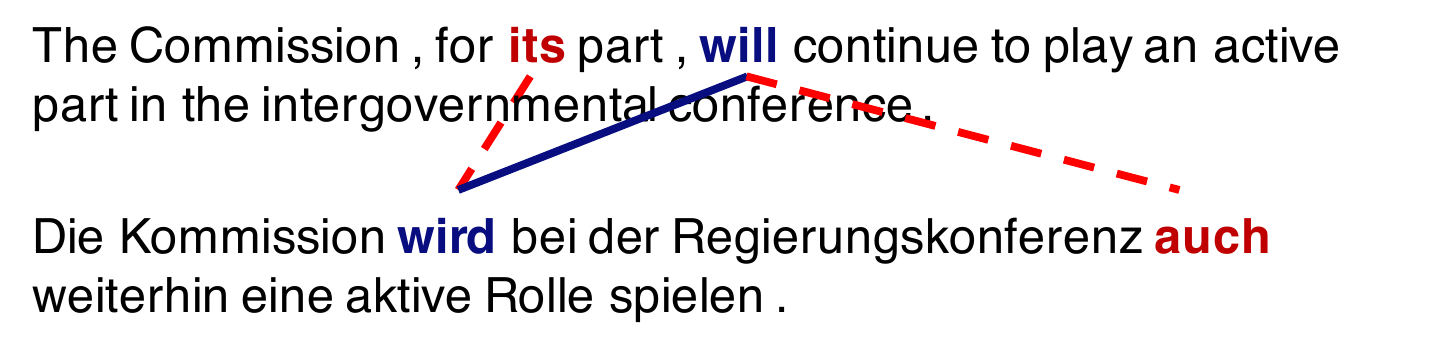}
	\caption{Example alignment of auxiliary verbs.
Same setting as in \tabref{pos_table}.
		Solid lines:  mBERT's alignment, identical to the gold standard. Dashed lines:  eflomal's incorrect alignment.}
	\figlabel{aux_example}
\end{figure}

To analyze the performance with respect to
different part-of-speech (POS) tags, the ENG-DEU gold
standard was tagged with the Stanza
toolkit \cite{qi2020stanza}.
We evaluate the alignment performance
for each POS tag by only considering the alignment edges
where at least one of their member words has this  tag.  \tabref{pos_table} shows results
for frequent POS tags.  Compared to eflomal,
mBERT aligns 
auxiliaries, pronouns and verbs better.
The relative position of
auxiliaries and verbs in German can diverge strongly  from
that in English because they occur at the end of the
sentence (verb-end position) in many clause types.
Positions of pronouns can also diverge due to a more
flexible word order in German. 
It is difficult for an
HMM-based aligner like eflomal to model such
high-distortion alignments, a property that has been found by prior work as well \cite{honeural}. In contrast, mBERT(Argmax) does
not use distortion information, so high distortion is not a
problem for it.

\figref{aux_example} gives an example for auxiliaries.
The gold alignment (``has'' -- ``hat'') is correctly
identified by mBERT (solid line).
Eflomal generates an incorrect  alignment
(``time'' -- ``hat''): the two words have about the same
relative position, indicating that distortion minimization
is the main reason for this incorrect alignment.
Analyzing all auxiliary alignment edges, the average absolute value of the 
distance between aligned words is $2.72$ for eflomal and
$3.22$ for mBERT. This indicates that eflomal is more
reluctant than mBERT to generate high-distortion alignments
and thus loses accuracy.

\section{Related Work}
\newcite{brown1993mathematics} introduced
the IBM models, the best known statistical word aligners.
More recent aligners, often based on IBM models,
include fast-align \cite{dyer2013simple}, Giza++
\cite{och03:asc} and
eflomal \cite{ostling2016efficient}. \cite{ostling2015bayesian} showed that Bayesian Alignment Models
perform well.
Neural network based extensions 
of these models have been considered \cite{ayan2005neuralign,honeural}.
All of these models are trained on
parallel text.
Our method instead aligns based on embeddings
that are induced from monolingual data only. We compare with prior methods and observe comparable performance.

Prior work on using learned representations for alignment
includes 
\cite{smadja1996translating,och03:asc} (Dice coefficient),
\cite{sabet2016improving} (incorporation of  embeddings
into IBM models),
\cite{legrand2016neural} (neural network alignment model)
and \cite{pourdamghani2018using} (embeddings are used to encourage words to align to similar words).
\citet{tamura2014recurrent} use recurrent neural networks to learn alignments. 
They use noise contrastive estimation
to avoid supervision. \citet{yang2013word} train a
neural network that uses pretrained word embeddings in the initial layer. 
All of this work requires parallel data. 
mBERT is used for word alignments in concurrent work: 
\newcite{libovicky2019language}
use the high quality of  mBERT alignments
as evidence for 
the ``language-neutrality'' of mBERT. \newcite{nagata2020supervised} phrase word alignment as crosslingual span prediction and finetune mBERT using gold alignments.

Attention in  NMT \cite{bahdanau2014neural}
is related to a notion of soft alignment, but
often deviates from conventional word alignments
\cite{ghader2017does,koehn2017six}. One difference is
that standard attention  does not have access to the
target word.
To address this,
\newcite{peter2017generating} tailor
attention matrices to obtain higher quality
alignments. \newcite{li2018target}'s and \newcite{zenkel2019adding}'s
models perform similarly to and \newcite{zenkel2020end} outperform
Giza++.
\citet{ding2019saliency} propose better decoding algorithms to 
deduce word alignments from NMT predictions.
\newcite{chen2016guided}, \newcite{mi2016supervised} and
\newcite{garg2019jointly} 
obtain alignments and translations
in a multitask setup. \newcite{garg2019jointly} find that operating at the
subword level can be beneficial for alignment
models. \citet{li2019word} propose two methods to extract alignments from 
NMT models, however they do not outperform fast-align.
\newcite{stengel2019discriminative} 
compute
similarity matrices
of encoder-decoder representations that are leveraged for 
word
alignments, together with supervised learning, which requires manually annotated alignment. 
We find our proposed methods to be competitive with these approaches. 
In contrast to our work, they all require parallel data.

\section{Conclusion}

We presented word aligners based on contextualized embeddings
that outperform in four and match the performance of 
state-of-the-art aligners in two language pairs; e.g., for ENG-DEU 
contextualized embeddings achieve
an alignment $F_1$ that is \ppnumpar \ percentage points higher than eflomal
trained on \numpar \ parallel sentences.
Further, we showed that alignments from 
static embeddings can be a viable alternative to statistical aligner when few parallel training 
data is available.
In contrast to all prior work our methods do not require parallel data 
for training at all.
With our proposed methods and extensions such as Match, Itermax and Null
it is easy to obtain higher precision or recall depending on the use case. 

Future work includes modeling fertility explicitly and investigating
how to incorporate parallel data into the proposed methods.

\section*{Acknowledgments}
	We gratefully
	acknowledge
	funding through a
	Zentrum Digitalisierung.Bayern 
	fellowship awarded to the second author. This work was supported
	by the European Research Council (\# 740516) and the German Federal Ministry of Education and
	Research (BMBF) under Grant No. 01IS18036A.
	The authors of this work take full responsibility for its content.
	We thank Matthias Huck, Jind\v{r}ich Libovick\'y, Alex
        Fraser and the anonymous reviewers for interesting discussions and valuable comments. 
        Thanks to Jind\v{r}ich for pointing out that mBERT can align
        mixed-language sentences as shown in \figref{firstpageexample}.

% \bibliography{acl2019}
\bibliographystyle{acl_natbib}

\appendix

\def\myhref#1{\href{#1}{#1}}

\section{Additional Non-central Results}
\seclabel{results}

\subsection{Comparison with Prior Work}

A more detailed version of Table 2 from the main paper that includes precision and recall 
and results on Itermax can be found in \tabref{mainsupp}.

\subsection{Rare Words}
\figref{rarewords1000k} shows the same as Figure 6 
from the main paper but now with a reference corpus of 100k/1000k instead of 1920k
parallel sentences. The main takeaways are similar.

\begin{table*}[t]
	\scriptsize
	\centering
	\def\tablesep{0.01cm}
	\begin{tabular}{
			@{\hspace{\tablesep}}c@{\hspace{\tablesep}}
			@{\hspace{\tablesep}}c@{\hspace{\tablesep}}|
			@{\hspace{\tablesep}}l@{\hspace{\tablesep}}||
			@{\hspace{\tablesep}}c@{\hspace{\tablesep}}
			@{\hspace{\tablesep}}c@{\hspace{\tablesep}}
			@{\hspace{\tablesep}}c@{\hspace{\tablesep}}
			@{\hspace{\tablesep}}c@{\hspace{\tablesep}}|
			@{\hspace{\tablesep}}c@{\hspace{\tablesep}}
			@{\hspace{\tablesep}}c@{\hspace{\tablesep}}
			@{\hspace{\tablesep}}c@{\hspace{\tablesep}}
			@{\hspace{\tablesep}}c@{\hspace{\tablesep}}|
			@{\hspace{\tablesep}}c@{\hspace{\tablesep}}
			@{\hspace{\tablesep}}c@{\hspace{\tablesep}}
			@{\hspace{\tablesep}}c@{\hspace{\tablesep}}
			@{\hspace{\tablesep}}c@{\hspace{\tablesep}}|
			@{\hspace{\tablesep}}c@{\hspace{\tablesep}}
			@{\hspace{\tablesep}}c@{\hspace{\tablesep}}
			@{\hspace{\tablesep}}c@{\hspace{\tablesep}}
			@{\hspace{\tablesep}}c@{\hspace{\tablesep}}|
			@{\hspace{\tablesep}}c@{\hspace{\tablesep}}
			@{\hspace{\tablesep}}c@{\hspace{\tablesep}}
			@{\hspace{\tablesep}}c@{\hspace{\tablesep}}
			@{\hspace{\tablesep}}c@{\hspace{\tablesep}}|
			@{\hspace{\tablesep}}c@{\hspace{\tablesep}}
			@{\hspace{\tablesep}}c@{\hspace{\tablesep}}
			@{\hspace{\tablesep}}c@{\hspace{\tablesep}}
			@{\hspace{\tablesep}}c@{\hspace{\tablesep}}}
		& & & \multicolumn{4}{c}{ ENG-CES } & \multicolumn{4}{c}{ ENG-DEU } & \multicolumn{4}{c}{ ENG-FAS } & \multicolumn{4}{c}{ ENG-FRA } & \multicolumn{4}{c}{ ENG-HIN } & \multicolumn{4}{c}{ ENG-RON } \\
		& & Method & Prec. & Rec. & $F_1$ & AER & Prec. & Rec. & $F_1$ & AER & Prec. & Rec. & $F_1$ & AER & Prec. & Rec. & $F_1$ & AER & Prec. & Rec. & $F_1$ & AER & Prec. & Rec. & $F_1$ & AER \\
		\midrule
		\midrule
\multirow{7}{*}{ \rotatebox{90}{Prior Work} }
& & \cite{ostling2015bayesian} Bayesian & & & & & & & & & & & & & .96 & .92 & \textbf{.94 }& \textbf{.06 }& .85 & .43 & .57 & .43 & .91 & .61 & \textbf{.73 }& \textbf{.27 }\\
& & \cite{ostling2015bayesian} Giza++ & & & & & & & & & & & & & \textbf{.98 }& .87 & .92 & .07 & .63 & .44 & .51 & .49 & .85 & .63 & .72 & .28 \\
& & \cite{legrand2016neural} Ensemble Method & .79 & .83 & .81 & .16 & & & & & & & & & .59 & .90 & .71 & .10 & & & & & & & &  \\
& & \cite{ostling2016efficient} efmaral & & & & & & & & & & & & & & & .93 & .08 & & & .53 & .47 & & & .72 & .28 \\
& & \cite{ostling2016efficient} fast-align & & & & & & & & & & & & & & & .86 & .15 & & & .33 & .67 & & & .68 & .33 \\
& & \cite{zenkel2019adding} Giza++ & & & & & & & & .21 & & & & & & & & \textbf{.06} & & & & & & & & .28\\
& & \cite{garg2019jointly} Multitask & & & & & & & & .20 & & & & & & & & .08 & & & & & & & &  \\
\midrule
\multirow{6}{*}{ \rotatebox{90}{Baselines} } & \multirow{3}{*}{ \rotatebox{90}{Word} } & fast-align/IBM2 & .71 & .81 & .76 & .25 & .70 & .73 & .71 & .29 & .60 & .54 & .57 & .43 & .81 & .93 & .86 & .15 & .34 & .33 & .34 & .66 & .69 & \textbf{.67 }& .68 & .33 \\
& & Giza++/IBM4 & .71 & .79 & .75 & .26 & .79 & .75 & .77 & .23 & .55 & .48 & .51 & .49 & .90 & .95 & .92 & .09 & .47 & .43 & .45 & .55 & .74 & .64 & .69 & .31 \\
& & eflomal & .84 & .86 & .85 & .15 & .80 & .75 & .77 & .23 & .68 & .55 & .61 & .39 & .91 & .94 & .93 & .08 & .61 & .44 & .51 & .49 & .81 & .63 & .71 & .29 \\
\cmidrule{2-27}
& \multirow{3}{*}{ \rotatebox{90}{Subword} } & fast-align/IBM2 & .72 & .84 & .78 & .23 & .67 & .74 & .71 & .30 & .60 & .56 & .58 & .42 & .80 & .92 & .85 & .16 & .39 & .37 & .38 & .62 & .69 & \textbf{.67 }& .68 & .32 \\
& & Giza++/IBM4 & .79 & .86 & .82 & .18 & .78 & .78 & .78 & .22 & .58 & .56 & .57 & .43 & .89 & .95 & .92 & .09 & .52 & .44 & .48 & .52 & .74 & .64 & .69 & .32 \\
& & eflomal & .80 & .88 & .84 & .17 & .74 & .78 & .76 & .24 & .66 & .60 & .63 & .37 & .88 & .95 & .91 & .09 & .58 & .47 & .52 & .48 & .78 & \textbf{.67 }& .72 & .28 \\
\midrule
\multirow{12}{*}{ \rotatebox{90}{This Work} } & \multirow{6}{*}{ \rotatebox{90}{Word} } & fastText - Itermax & .74 & .69 & .72 & .29 & .69 & .56 & .62 & .38 & .63 & .45 & .53 & .48 & .74 & .78 & .76 & .24 & .63 & .42 & .51 & .49 & .64 & .40 & .50 & .51 \\
& & mBERT[8] - Itermax & .87 & .87 & \textbf{.87 }& .14 & .85 & .77 & \textbf{.81 }& \textbf{.19 }& .80 & .63 & .70 & .30 & .91 & .95 & .93 & .08 & .75 & .47 & .58 & .43 & .82 & .58 & .68 & .32 \\
& & XLM-R[8] - Itermax & .89 & .85 & \textbf{.87 }& \textbf{.13 }& .86 & .73 & .79 & .21 & .84 & .63 & \textbf{.72 }& \textbf{.28 }& .91 & .93 & .92 & .08 & .79 & .49 & .61 & \textbf{.39} & .87 & .61 & .71 & .29 \\
& & fastText - Argmax & .86 & .59 & .70 & .30 & .81 & .48 & .60 & .40 & .75 & .38 & .50 & .50 & .85 & .71 & .77 & .22 & .75 & .36 & .49 & .52 & .77 & .34 & .47 & .53 \\
& & mBERT[8] - Argmax & .95 & .80 & \textbf{.87 }& \textbf{.13 }& .92 & .69 & .79 & .21 & .88 & .54 & .67 & .33 & .97 & .91 & \textbf{.94 } & \textbf{.06 }& .84 & .39 & .54 & .47 & .90 & .50 & .64 & .36 \\
& & XLM-R[8] - Argmax & \textbf{.96 }& .80 & \textbf{.87 }& \textbf{.13 }& \textbf{.93 }& .68 & .79 & .22 & \textbf{.91 }& .57 & .70 & .30 & .96 & .91 & .93 & \textbf{.06 }& \textbf{.88 }& .45 & .59 & .41 & \textbf{.94 }& .56 & .70 & .30 \\
\cmidrule{2-27}
& \multirow{6}{*}{ \rotatebox{90}{Subword} } & fastText - Itermax & .61 & .57 & .59 & .41 & .63 & .54 & .58 & .42 & .20 & .07 & .11 & .90 & .70 & .76 & .73 & .28 & .14 & .05 & .07 & .93 & .56 & .38 & .45 & .55 \\
& & mBERT[8] - Itermax & .84 & \textbf{.89 }& .86 & .14 & .83 & \textbf{.80 }& \textbf{.81 }& \textbf{.19 }& .76 & .65 & .70 & .30 & .91 & \textbf{.96 }& .93 & .08 & .71 & .49 & .58 & .42 & .79 & .62 & .69 & .31 \\
& & XLM-R[8] - Itermax & .84 & \textbf{.89} & .86 & .14 & .83 & .78 & .80 & .20 & .79 & \textbf{.67 }& \textbf{.72 }& \textbf{.28 }& .89 & .94 & .92 & .09 & .75 & \textbf{.52 }& \textbf{.62 }& \textbf{.39 }& .83 & .64 & .72& .28 \\
& & fastText - Argmax & .72 & .48 & .58 & .42 & .75 & .45 & .56 & .44 & .27 & .06 & .09 & .91 & .80 & .67 & .73 & .26 & .14 & .02 & .04 & .96 & .67 & .31 & .43 & .58 \\
& & mBERT[8] - Argmax & .92 & .81 & .86 & .14 & .92 & .72 & \textbf{.81 }& \textbf{.19 }& .85 & .56 & .67 & .33 & .96 & .92 & \textbf{.94 }& \textbf{.06 }& .81 & .41 & .55 & .45 & .88 & .51 & .65 & .35 \\
& & XLM-R[8] - Argmax & .92 & .83 & \textbf{.87 }& \textbf{.13 }& .92 & .72 & \textbf{.81 }& \textbf{.19 }& .87 & .59 & .71 & .30 & .95 & .91 & .93 & .07 & .86 & .47 & .61& \textbf{.39} & .91 & .59 & .71 & .29 \\
	\end{tabular}
	\caption{Comparison of word and
		subword levels. Best overall result per column in bold.}
	\tablabel{mainsupp}
\end{table*}

\begin{table}[t]
	\tiny
	\centering
	\def\postprocsep{0.04cm}
	\begin{tabular}{@{\hspace{\postprocsep}}c@{\hspace{\postprocsep}}|@{\hspace{\postprocsep}}l@{\hspace{\postprocsep}}||@{\hspace{\postprocsep}}c@{\hspace{\postprocsep}}@{\hspace{\postprocsep}}c@{\hspace{\postprocsep}}@{\hspace{\postprocsep}}c@{\hspace{\postprocsep}}@{\hspace{\postprocsep}}c@{\hspace{\postprocsep}}|@{\hspace{\postprocsep}}c@{\hspace{\postprocsep}}@{\hspace{\postprocsep}}c@{\hspace{\postprocsep}}@{\hspace{\postprocsep}}c@{\hspace{\postprocsep}}@{\hspace{\postprocsep}}c@{\hspace{\postprocsep}}|@{\hspace{\postprocsep}}c@{\hspace{\postprocsep}}@{\hspace{\postprocsep}}c@{\hspace{\postprocsep}}@{\hspace{\postprocsep}}c@{\hspace{\postprocsep}}@{\hspace{\postprocsep}}c@{\hspace{\postprocsep}}}
		& & \multicolumn{4}{c}{ ENG-DEU } & \multicolumn{4}{c}{ ENG-CES } & \multicolumn{4}{c}{ ENG-HIN } \\
		Emb. & Method & Prec. & Rec. & $F_1$ & AER & Prec. & Rec. & $F_1$ & AER & Prec. & Rec. & $F_1$ & AER \\
		\midrule
		\midrule
		\multirow{9}{*}{ \rotatebox{90}{fastText} } & Argmax & .75 & .45 & .56 & .44 & .72 & .48 & .58 & .42 & .14 & .02 & .04 & .96 \\
		& \emph{+Dist} & \textbf{.79 }& \textbf{.51 }& \textbf{.62 }& \textbf{.38 }& \textbf{.77 }& \textbf{.58 }& \textbf{.66 }& \textbf{.34 }& \textbf{.16 }& \textbf{.04 }& \textbf{.06 }& \textbf{.94 }\\
		& \emph{+Null} & .76 & .43 & .55 & .45 & .74 & .47 & .57 & .42 & .14 & .02 & .04 & .96 \\
		\cmidrule{2-14}
		& Itermax & .63 & .54 & .58 & .42 & .61 & .57 & .59 & .41 & .14 & .05 & .07 & .93 \\
		& \emph{+Dist} & \textbf{.67 }& \textbf{.60 }& \textbf{.64 }& \textbf{.36 }& \textbf{.63 }& \textbf{.66 }& \textbf{.65 }& \textbf{.36 }& \textbf{.15 }& \textbf{.07 }& \textbf{.09 }& \textbf{.91 }\\
		& \emph{+Null} & .64 & .52 & .57 & .43 & .62 & .56 & .59 & .41 & .14 & .04 & .07 & .93 \\
		\cmidrule{2-14}
		& Match & .51 & .58 & .54 & .46 & .44 & .61 & .52 & .49 & \textbf{.10 }& .08 & \textbf{.09 }& \textbf{.91 }\\
		& \emph{+Dist} & \textbf{.59 }& \textbf{.66 }& \textbf{.62 }& \textbf{.38 }& \textbf{.54 }& \textbf{.71 }& \textbf{.61 }& \textbf{.39 }& \textbf{.10 }& \textbf{.09 }& \textbf{.09 }& \textbf{.91 }\\
		& \emph{+Null} & .52 & .57 & .54 & .46 & .46 & .60 & .52 & .48 & \textbf{.10 }& .08 & \textbf{.09 }& \textbf{.91 }\\
		\midrule
		\midrule
		\multirow{9}{*}{ \rotatebox{90}{mBERT[8]} } & Argmax & .92 & \textbf{.72 }& \textbf{.81 }& \textbf{.19 }& \textbf{.92 }& \textbf{.81 }& \textbf{.86 }& \textbf{.14 }& .81 & \textbf{.41 }& \textbf{.55 }& \textbf{.45 }\\
		& \emph{+Dist} & .90 & .70 & .79 & .21 & .91 & .80 & .85 & .15 & .65 & .30 & .41 & .59 \\
		& \emph{+Null} & \textbf{.93 }& .70 & .80 & .20 & \textbf{.92 }& .78 & .85 & .15 & \textbf{.82 }& .40 & .54 & .47 \\
		\cmidrule{2-14}
		& Itermax & .83 & \textbf{.80 }& \textbf{.81 }& \textbf{.19 }& .84 & \textbf{.89 }& \textbf{.86 }& \textbf{.14 }& .71 & \textbf{.49 }& \textbf{.58 }& \textbf{.42 }\\
		& \emph{+Dist} & .81 & .77 & .79 & .21 & .82 & .87 & .84 & .16 & .53 & .35 & .42 & .58 \\
		& \emph{+Null} & \textbf{.85 }& .77 & \textbf{.81 }& .20 & \textbf{.84 }& .86 & .85 & .15 & \textbf{.72 }& .47 & .57 & .43 \\
		\cmidrule{2-14}
		& Match & .75 & \textbf{.80 }& \textbf{.78 }& \textbf{.23 }& .76 & \textbf{.90 }& \textbf{.82 }& \textbf{.18 }& .64 & \textbf{.52 }& \textbf{.58 }& \textbf{.43 }\\
		& \emph{+Dist} & .72 & .77 & .75 & .26 & .74 & .88 & .80 & .20 & .45 & .37 & .40 & .60 \\
		& \emph{+Null} & \textbf{.77 }& .78 & \textbf{.78 }& \textbf{.23 }& \textbf{.77 }& .88 & \textbf{.82 }& .19 & \textbf{.65 }& .51 & .57 & \textbf{.43 }\\
	\end{tabular}
	\caption{Comparison of methods for inducing alignments from similarity matrices. All results are subword-level. Best result per embedding type across columns in bold.}
	\tablabel{postprocsubword}
\end{table}

\begin{figure}[t]
	\centering
	\includegraphics[width=0.9\linewidth]{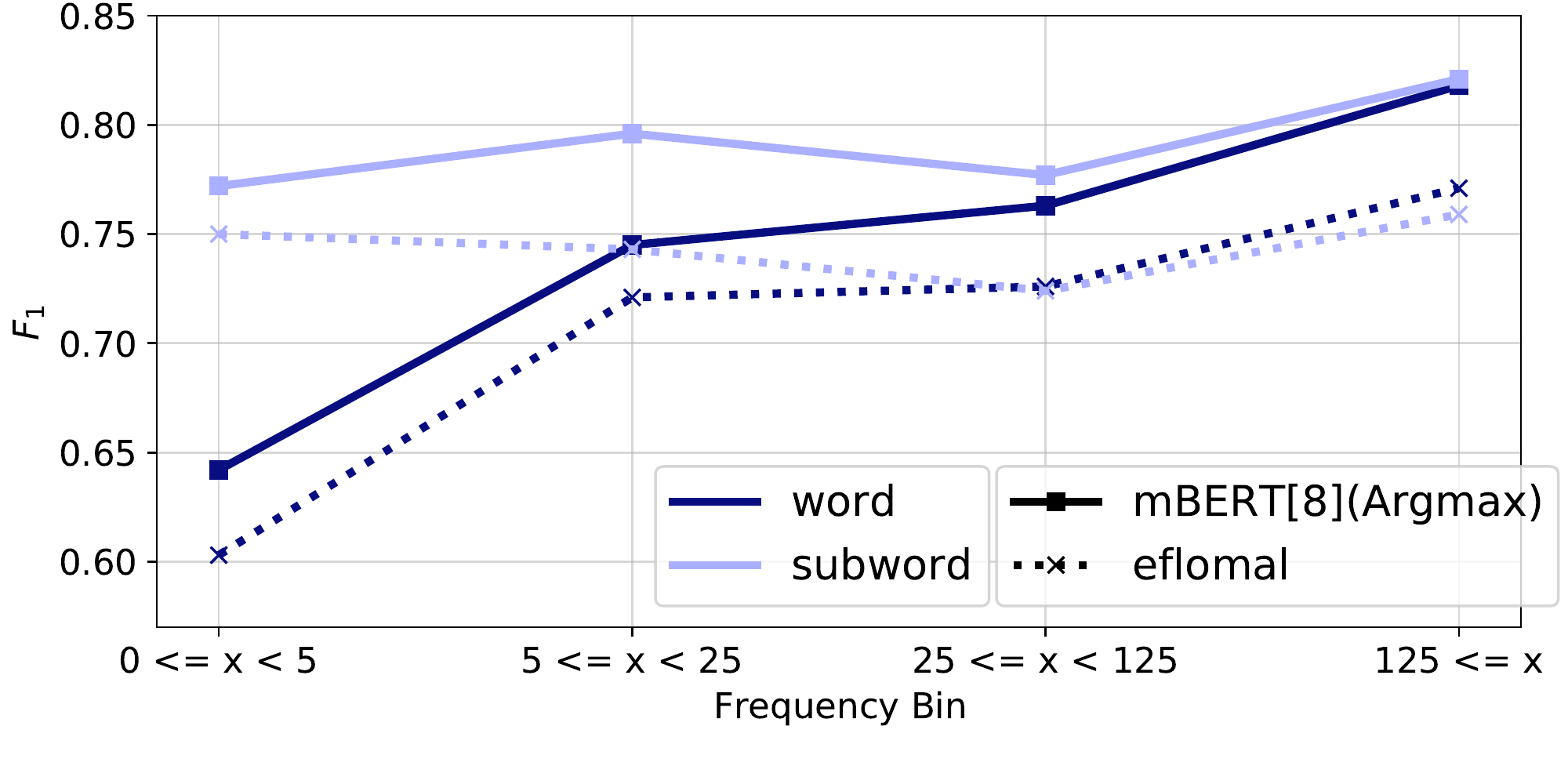}
	\includegraphics[width=0.9\linewidth]{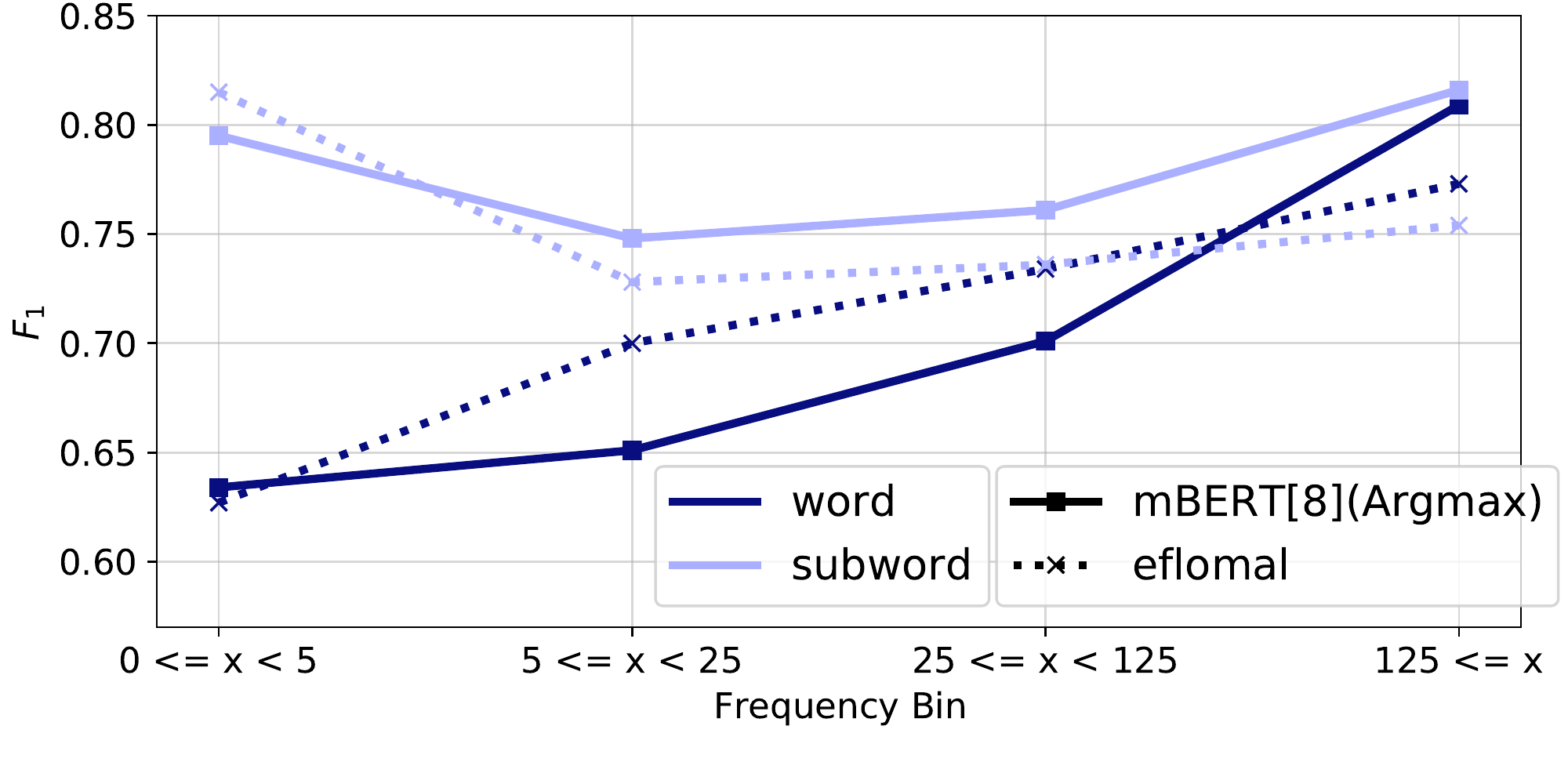}
	\caption{Results for different frequency bins. An edge in $S$, $P$, or $A$ 
		is attributed to exactly one bin based on the minimum frequency of the involved words (denoted by $x$). 
		Top: Eflomal trained and frequencies computed on 100k parallel sentences. Bottom: 1000k parallel sentences.
	}
	\figlabel{rarewords1000k}
\end{figure}

\subsection{Symmetrization}
For asymmetric alignments different symmetrization methods exist.
\citet{dyer2013simple} provide an overview and implementation
(fast-align) for these methods, which we use. 
We compare intersection and grow-diag-final-and (GDFA) in \tabref{symm}. 
In terms of F1, GDFA performs better (Intersection wins four times, GDFA eleven times, three ties). 
As expected, Intersection yields higher precision while GDFA yields higher recall.
Thus intersection is preferable for tasks like annotation projection, whereas GDFA is typically used in statistical machine translation.

\begin{table*}[t]
	\scriptsize
	\centering
	\def\tablesep{0.02cm}
	\begin{tabular}{
			@{\hspace{\tablesep}}c@{\hspace{\tablesep}}|
			@{\hspace{\tablesep}}l@{\hspace{\tablesep}}||
			@{\hspace{\tablesep}}c@{\hspace{\tablesep}}
			@{\hspace{\tablesep}}c@{\hspace{\tablesep}}
			@{\hspace{\tablesep}}c@{\hspace{\tablesep}}
			@{\hspace{\tablesep}}c@{\hspace{\tablesep}}|
			@{\hspace{\tablesep}}c@{\hspace{\tablesep}}
			@{\hspace{\tablesep}}c@{\hspace{\tablesep}}
			@{\hspace{\tablesep}}c@{\hspace{\tablesep}}
			@{\hspace{\tablesep}}c@{\hspace{\tablesep}}|
			@{\hspace{\tablesep}}c@{\hspace{\tablesep}}
			@{\hspace{\tablesep}}c@{\hspace{\tablesep}}
			@{\hspace{\tablesep}}c@{\hspace{\tablesep}}
			@{\hspace{\tablesep}}c@{\hspace{\tablesep}}|
			@{\hspace{\tablesep}}c@{\hspace{\tablesep}}
			@{\hspace{\tablesep}}c@{\hspace{\tablesep}}
			@{\hspace{\tablesep}}c@{\hspace{\tablesep}}
			@{\hspace{\tablesep}}c@{\hspace{\tablesep}}|
			@{\hspace{\tablesep}}c@{\hspace{\tablesep}}
			@{\hspace{\tablesep}}c@{\hspace{\tablesep}}
			@{\hspace{\tablesep}}c@{\hspace{\tablesep}}
			@{\hspace{\tablesep}}c@{\hspace{\tablesep}}|
			@{\hspace{\tablesep}}c@{\hspace{\tablesep}}
			@{\hspace{\tablesep}}c@{\hspace{\tablesep}}
			@{\hspace{\tablesep}}c@{\hspace{\tablesep}}
			@{\hspace{\tablesep}}c@{\hspace{\tablesep}}}
		& & \multicolumn{4}{c}{ ENG-CES } & \multicolumn{4}{c}{ ENG-DEU } & \multicolumn{4}{c}{ ENG-FAS } & \multicolumn{4}{c}{ ENG-FRA } & \multicolumn{4}{c}{ ENG-HIN } & \multicolumn{4}{c}{ ENG-RON } \\
		Method & Symm. & Prec. & Rec. & $F_1$ & AER & Prec. & Rec. & $F_1$ & AER & Prec. & Rec. & $F_1$ & AER & Prec. & Rec. & $F_1$ & AER & Prec. & Rec. & $F_1$ & AER & Prec. & Rec. & $F_1$ & AER \\
		\midrule
		\midrule
		\multirow{2}{1cm}{eflomal} & Inters. & \textbf{.95 }& .79 & \textbf{.86 }& \textbf{.14 }& \textbf{.91 }& .66 & .76 & .24 & \textbf{.88 }& .43 & .58 & .42 & \textbf{.96 }& .90 & \textbf{.93 }& \textbf{.07 }& \textbf{.81 }& .37 & \textbf{.51 }& \textbf{.49 }& \textbf{.91 }& .56 & .70 & .31 \\
		& GDFA & .84 & \textbf{.86 }& .85 & .15 & .80 & \textbf{.75 }& \textbf{.77 }& \textbf{.23 }& .68 & \textbf{.55 }& \textbf{.61 }& \textbf{.39 }& .91 & \textbf{.94 }& \textbf{.93 }& .08 & .61 & \textbf{.44 }& \textbf{.51 }& \textbf{.49 }& .81 & \textbf{.63 }& \textbf{.71 }& \textbf{.29 }\\
		\midrule
		\multirow{2}{1cm}{fast-align} & Inters. & \textbf{.89 }& .69 & \textbf{.78 }& \textbf{.22 }& \textbf{.87 }& .60 & \textbf{.71 }& \textbf{.29 }& \textbf{.78 }& .43 & .55 & .45 & \textbf{.93 }& .84 & \textbf{.88 }& \textbf{.11 }& \textbf{.55 }& .22 & .31 & .69 & \textbf{.89 }& .50 & .64 & .36 \\
		& GDFA & .71 & \textbf{.81 }& .76 & .25 & .70 & \textbf{.73 }& \textbf{.71 }& \textbf{.29 }& .60 & \textbf{.54 }& \textbf{.57 }& \textbf{.43 }& .81 & \textbf{.93 }& .86 & .15 & .34 & \textbf{.33 }& \textbf{.34 }& \textbf{.66 }& .69 & \textbf{.67 }& \textbf{.68 }& \textbf{.33 }\\
		\midrule
		\multirow{2}{1cm}{GIZA++} & Inters. & \textbf{.95 }& .60 & .74 & \textbf{.26 }& \textbf{.92 }& .62 & .74 & .26 & \textbf{.89 }& .26 & .40 & .60 & \textbf{.97 }& .89 & \textbf{.93 }& \textbf{.06 }& \textbf{.82 }& .25 & .38 & .62 & \textbf{.95 }& .47 & .63 & .37 \\
		& GDFA & .71 & \textbf{.79 }& \textbf{.75 }& \textbf{.26 }& .79 & \textbf{.75 }& \textbf{.77 }& \textbf{.23 }& .55 & \textbf{.48 }& \textbf{.51 }& \textbf{.49 }& .90 & \textbf{.95 }& .92 & .09 & .47 & \textbf{.43 }& \textbf{.45 }& \textbf{.55 }& .74 & \textbf{.64 }& \textbf{.69 }& \textbf{.31 }\\
	\end{tabular}
	\caption{Comparison of symmetrization methods at the word level. Best result across rows per method in bold.}
	\tablabel{symm}
\end{table*}

\subsection{Alignment Examples for Different Methods}

We show examples in \figref{examples1}, 
\figref{examples3}, 
\figref{examples4}, 
and \figref{examples5}.
They provide an overview how the methods actually affect results.

\section{Hyperparameters}
\seclabel{hyper}

\subsection{Overview}
We provide a list of customized hyperparameters used in our computations in \tabref{hyperparams}.
There are three options how we came up with the hyperparameters: 
a) We simply used default values of 3rd party software. 
b) We chose an arbitrary value. Usually we fell back to well-established and rather conventional values (e.g., embedding dimension 300 for static embeddings).
c) We defined a reasonable but arbitrary range, out of which we selected the best value using grid search. 
\tabref{hyperparams} lists the final values we used as well as how we came up with the specific value. 
For option c) the corresponding analyses are in Figure 4 and Table 3 in the main paper as well as 
in \secref{nulldist} in this supplementary material.

\begin{table*}[t]
	\scriptsize
	\centering
	\begin{tabular}{c||lp{9cm}}
		\textbf{System} & \textbf{Parameter} & \textbf{Value} \\
		\midrule
		\multirow{4}{50pt}{fastText}
		&Version &0.9.1 \\
		&Code URL & https://github.com/facebookresearch/fastText/archive/v0.9.1.zip\\
		&Downloaded on & 11.11.2019 \\
		&Embedding Dimension&300\\
		\midrule
		\multirow{2}{50pt}{mBERT,XLM-R}
		&Code: Huggingface Transformer& Version 2.3.1 \\
		&Maximum Sequence Length&128\\
		\midrule
		\multirow{3}{50pt}{fastalign}
		&Code URL&https://github.com/clab/fast\_align\\
		&Git Hash & 7c2bbca3d5d61ba4b0f634f098c4fcf63c1373e1\\
		&Flags&-d -o -v\\
		\midrule
		\multirow{3}{50pt}{eflomal}
		&Code URL&https://github.com/robertostling/eflomal\\
		&Git Hash &9ef1ace1929c7687a4817ec6f75f47ee684f9aff \\
		&Flags&--model 3\\
		\midrule
		\multirow{3}{50pt}{GIZA++}
		&Code URL&http://web.archive.org/web/20100221051856/http://code.google.com/p/giza-pp\\
		&Version &1.0.3\\
		&Iterations& 5 iter. HMM, 5 iter. Model 1, 5 iter. Model3, 5 iter. Model 4 (DEFAULT) \\
		&p0&0.98\\
		\midrule
		\multirow{3}{50pt}{Vecmap}
		&Code URL& https://github.com/artetxem/vecmap.git\\
		&Git Hash& b82246f6c249633039f67fa6156e51d852bd73a3\\
		&Manual Vocabulary Cutoff&500000\\
		\midrule
		\multirow{1}{50pt}{Distortion Ext.}  & $\kappa$ & 0.5 (chosen ouf of $[0.0, 0.1, \dots, 1.0]$ by grid search, criterion: $F_1$)\\
		\midrule
		\multirow{1}{50pt}{Null Extension}  &$\tau$& 95th percentile of similarity distribution of aligned edges (chosen out of [80, 90, 95, 98, 99, 99.5] by grid search, criterion: $F_1$)\\
		\midrule
		\multirow{1}{50pt}{Argmax}  &Layer& 8 (for mBERT and XLM-R, chosen out of $[0, 1, \dots, 12]$ by grid search, criterion: $F_1$ )\\
		\midrule
		\multirow{2}{50pt}{Vecmap} &$\alpha$& 0.9 (chosen out of [0.9, 0.95, 1] by grid search, criterion: $F_1$)\\
		&Iterations $n_{\max}$& 2 (chosen out of [1,2,3] by grid search, criterion: $F_1$)\\
	\end{tabular}
	\caption{Overview on hyperparameters. We only list parameters where we do \textbf{not} use default values.
		Shown are the values which we use unless specifically indicated otherwise. }
	\tablabel{hyperparams}
\end{table*}

\begin{table*}[h]
	\centering
	\scriptsize
	\begin{tabular}{l||rrl}
		Lang. & Name & Description & Link \\
		\midrule
		ENG-CES & \cite{marecek:2008} & Gold Alignment & \myhref{http://ufal.mff.cuni.cz/czech-english-manual-word-alignment} \\
		ENG-DEU & EuroParl-based & Gold Alignment & \myhref{www-i6.informatik.rwth-aachen.de/goldAlignment/}\\
		ENG-FAS & \cite{tavakoli2014phrase} & Gold Alignment & \myhref{http://eceold.ut.ac.ir/en/node/940}\\
		ENG-FRA &  WPT2003, \cite{och2000improved},& Gold Alignment & \myhref{http://web.eecs.umich.edu/~mihalcea/wpt/}\\
		ENG-HIN &  WPT2005 & Gold Alignment & \myhref{http://web.eecs.umich.edu/~mihalcea/wpt05/} \\
		ENG-RON &  WPT2005 \cite{mihalcea2003evaluation} & Gold Alignment & \myhref{http://web.eecs.umich.edu/~mihalcea/wpt05/} \\
		\midrule
		ENG-CES & EuroParl \cite{koehn2005europarl}& Parallel Data & \myhref{https://www.statmt.org/europarl/}\\
		ENG-DEU & EuroParl \cite{koehn2005europarl} & Parallel Data & \myhref{https://www.statmt.org/europarl/}\\
		ENG-DEU & ParaCrawl  & Parallel Data & \myhref{https://paracrawl.eu/}\\
		ENG-FAS & TEP \cite{pilevar2011tep} & Parallel Data & \myhref{http://opus.nlpl.eu/TEP.php} \\
		ENG-FRA &  Hansards \cite{germann2001aligned}  & Parallel Data & \myhref{https://www.isi.edu/natural-language/download/hansard/index.html}\\
		ENG-HIN & Emille \cite{mcenery2000emille}  & Parallel Data& \myhref{http://web.eecs.umich.edu/\~mihalcea/wpt05/} \\
		ENG-RON &  Constitution, Newspaper& Parallel Data&  \myhref{http://web.eecs.umich.edu/~mihalcea/wpt05/}\\
		\midrule
		All langs. & Wikipedia (downloaded October 2019) & Monolingual Text & \myhref{download.wikimedia.org/[X]wiki/latest/[X]wiki-latest-pages-articles.xml.bz2} \\
	\end{tabular}
	\caption{Overview of datasets. ``Lang.'' uses ISO 639-3 language codes.\tablabel{data}}
\end{table*}

\subsection{Null and Distortion Extensions}
\seclabel{nulldist}
\begin{figure}[t]
	\includegraphics[width=0.95\linewidth]{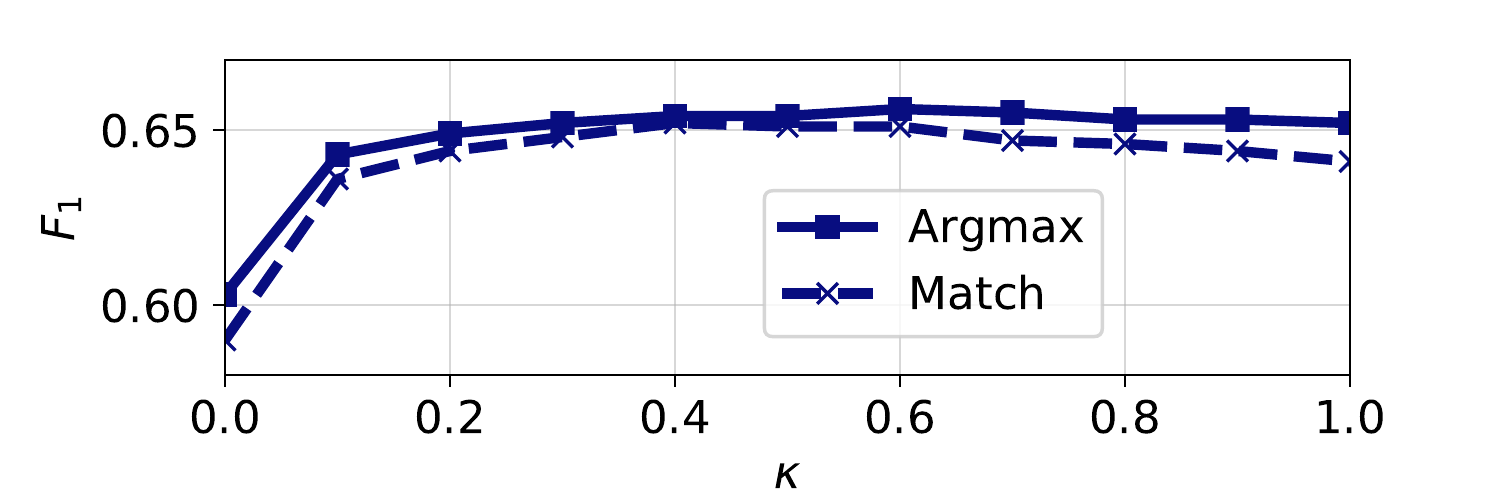}
	\includegraphics[width=0.95\linewidth]{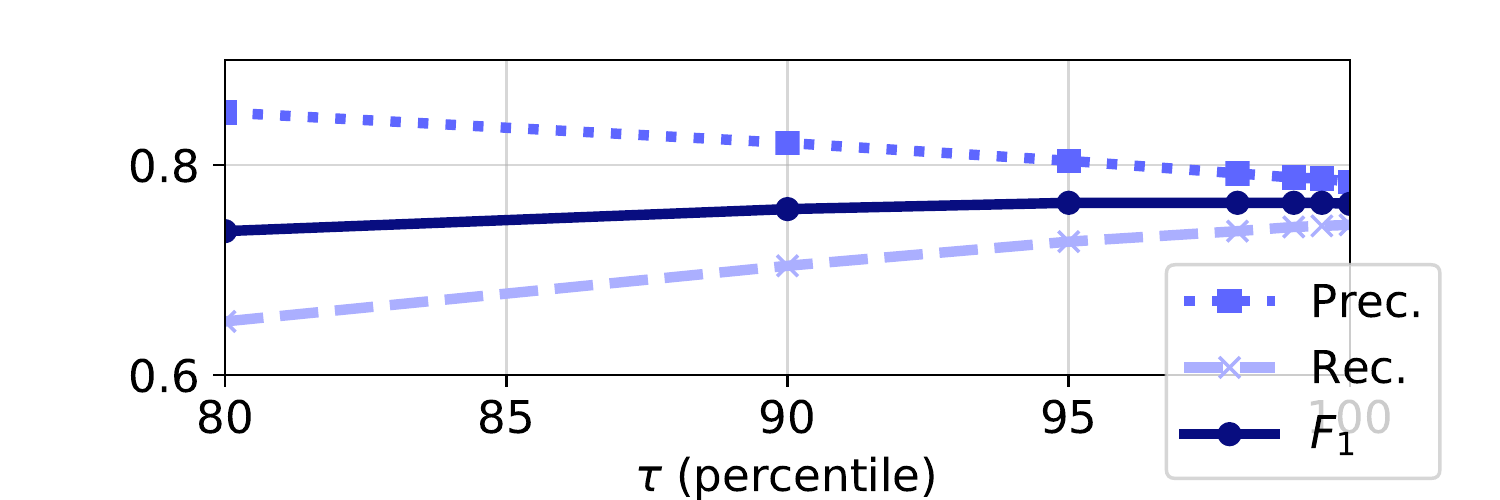}
	\caption{Top: F1 for ENG-DEU with fastText at word-level for different values of $\kappa$.
		Bottom: Performance for ENG-DEU with mBERT[8]
		(Match) at word-level when setting the value of $\tau$ to
		different percentiles. $\tau$ can be used for
		trading precision against recall. $F_1$ remains stable
		although it  decreases slightly when assigning $\tau$ the value
		of a smaller percentile (e.g., 80).}
	\figlabel{kapptau}
\end{figure}

In \figref{kapptau}
we plot the performance for different values of $\kappa$. We observe 
that introducing distortion indeed helps (i.e., $\kappa > 0$) but the actual 
value is not decisive for performance. This is rather intuitive, as a small 
adjustment to the similarities is sufficient while larger adjustments 
do not necessarily change the argmax or the optimal point in the matching algorithm. We choose
$\kappa =0.5$. 

For $\tau$ in null-word extension, we plot precision, recall and $F_1$ in \figref{kapptau} when assigning $\tau$ different percentile values. 
Note that values for $\tau$ depend on the similarity distribution of all aligned edges.  
As expected, when using the 100th percentile
no edges are removed and thus the performance is not changed compared to not having a null-word extension.
When decreasing the value of $\tau$ the precision increases and recall goes down, while $F_1$ remains stable.
We use the 95th percentile for $\tau$ .

\section{Reproducibility Information}
\seclabel{repro}

\subsection{Computing Infrastructures, Runtimes, Number of Parameters}
We did all computations on up to 48 cores of Intel(R) Xeon(R) CPU E7-8857 v2 with 1TB memory and a single
GeForce GTX 1080 GPU with 8GB memory. 

Runtimes for aligning 500 parallel sentences on ENG-DEU are reported in \tabref{runtimes}. mBERT and XLM-R computations are done on the GPU.
Note that fast-align, GIZA++ and eflomal usually need to be trained on much more parallel data
to achieve better performance: this increases their runtime.

All our proposed methods are \textbf{parameter-free}. If we consider the parameters of the pretrained language models
and pretrained embeddings then fastText has around 1 billion parameters (up to 500k words per language, 7 languages and embedding dimension 300), 
mBERT has 172 million, XLM-R 270 million parameters.
\begin{table}[H]
	\footnotesize
	\centering
	\begin{tabular}{l|r}
		Method & Runtime[s] \\
		\midrule
		fast-align & 4  \\
		GIZA++ & 18 \\
		eflomal & 5  \\
		mBERT[8] - Argmax &  15 \\
		XLM-R[8] - Argmax & 22 \\
	\end{tabular}
	\caption{Runtime (average across 5 runs) in seconds for each method to align 500 parallel sentences. \tablabel{runtimes}}
\end{table}

\subsection{Data}

\tabref{data} provides download links to all data used.

\begin{figure}[t]
	\centering
	\includegraphics[width=0.95\linewidth]{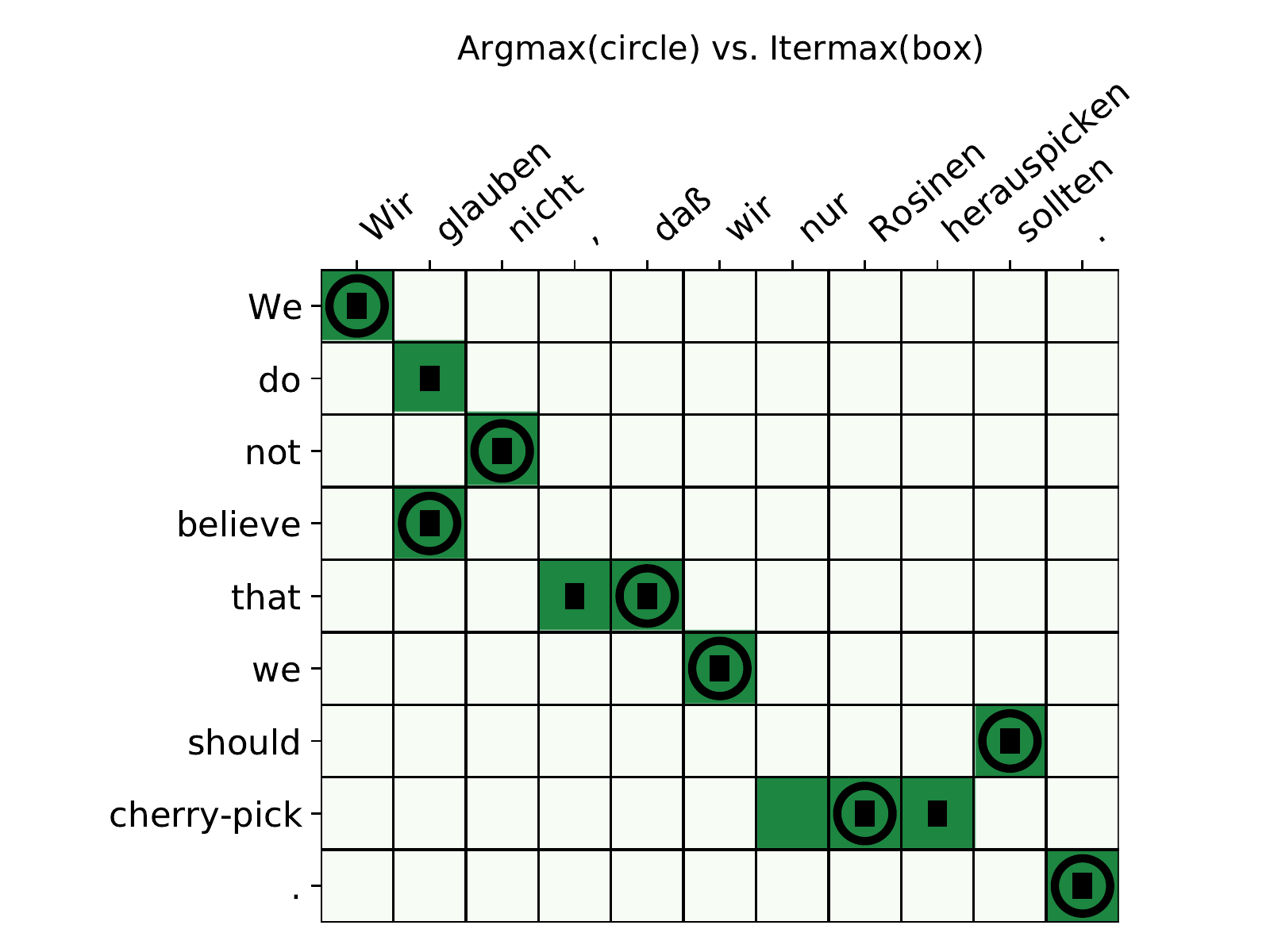}
	\includegraphics[width=0.95\linewidth]{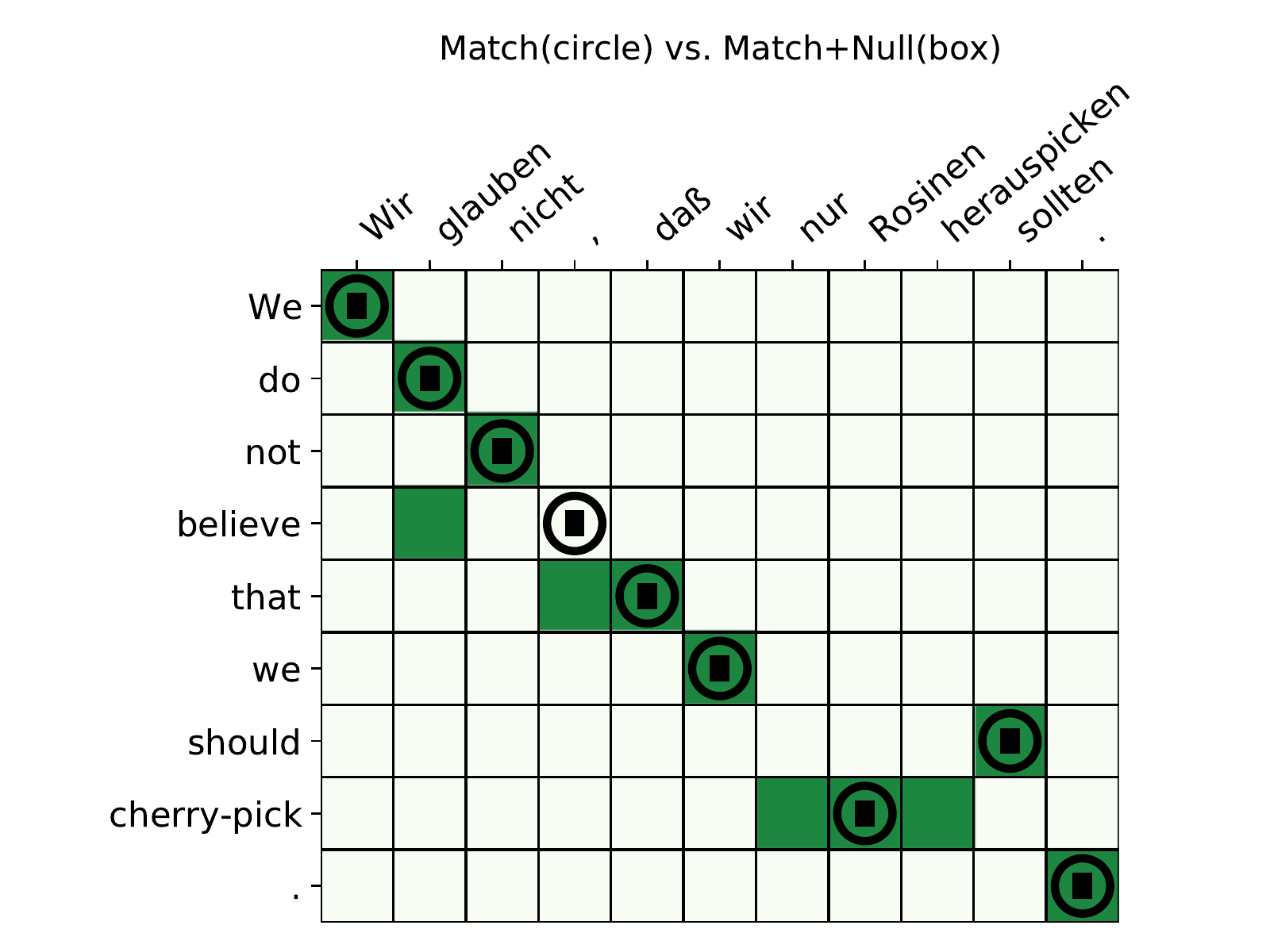}
	\includegraphics[width=0.95\linewidth]{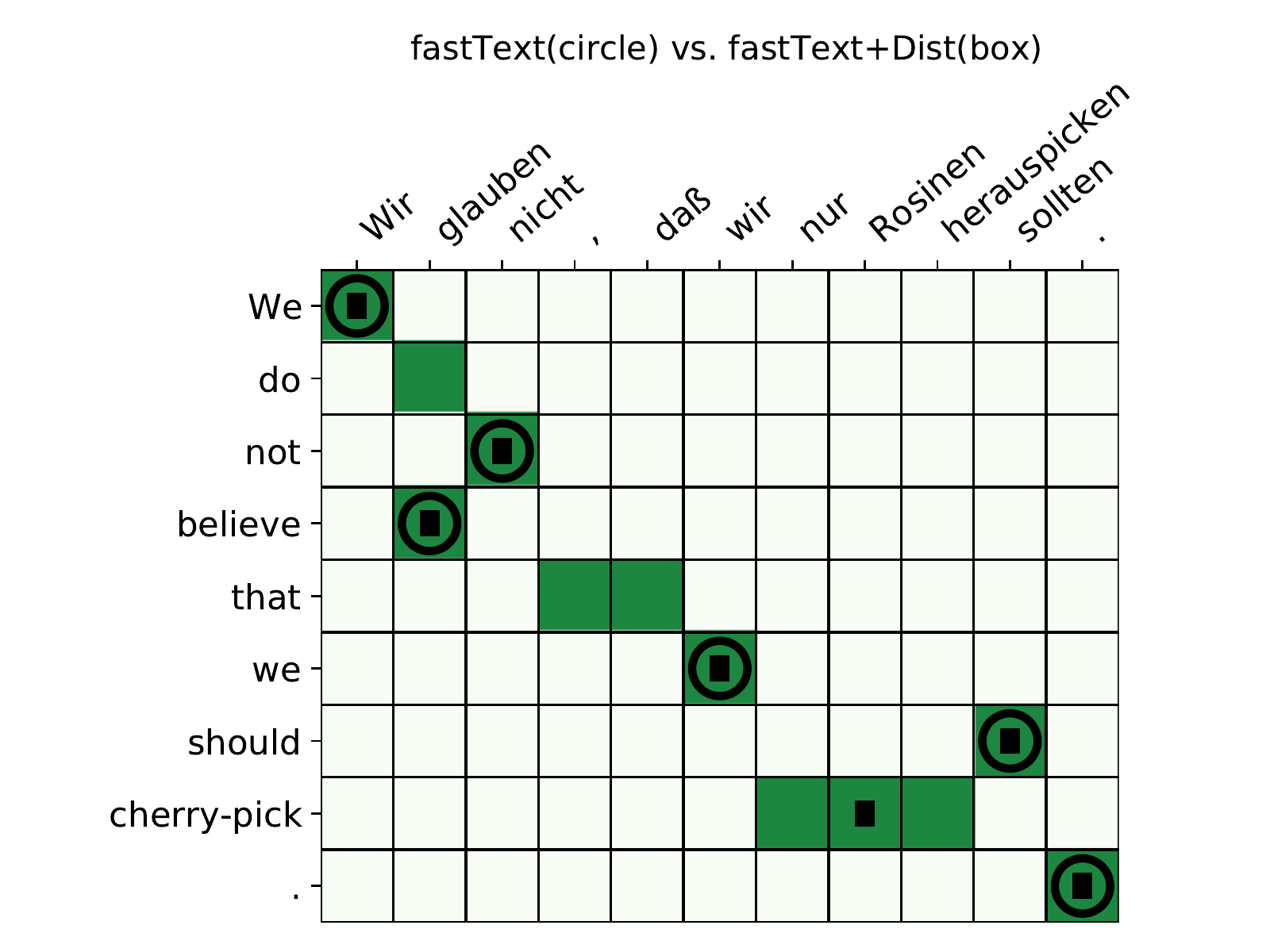}
	\caption{Comparison of alignment methods. Dark/light green: sure/possible edges in the gold standard. 
		Circles are alignments from the first
		mentioned method in the subfigure title, boxes alignments from the second method.}
	\figlabel{examples1}
\end{figure}
\begin{figure}[t]
	\centering
	\includegraphics[width=0.95\linewidth]{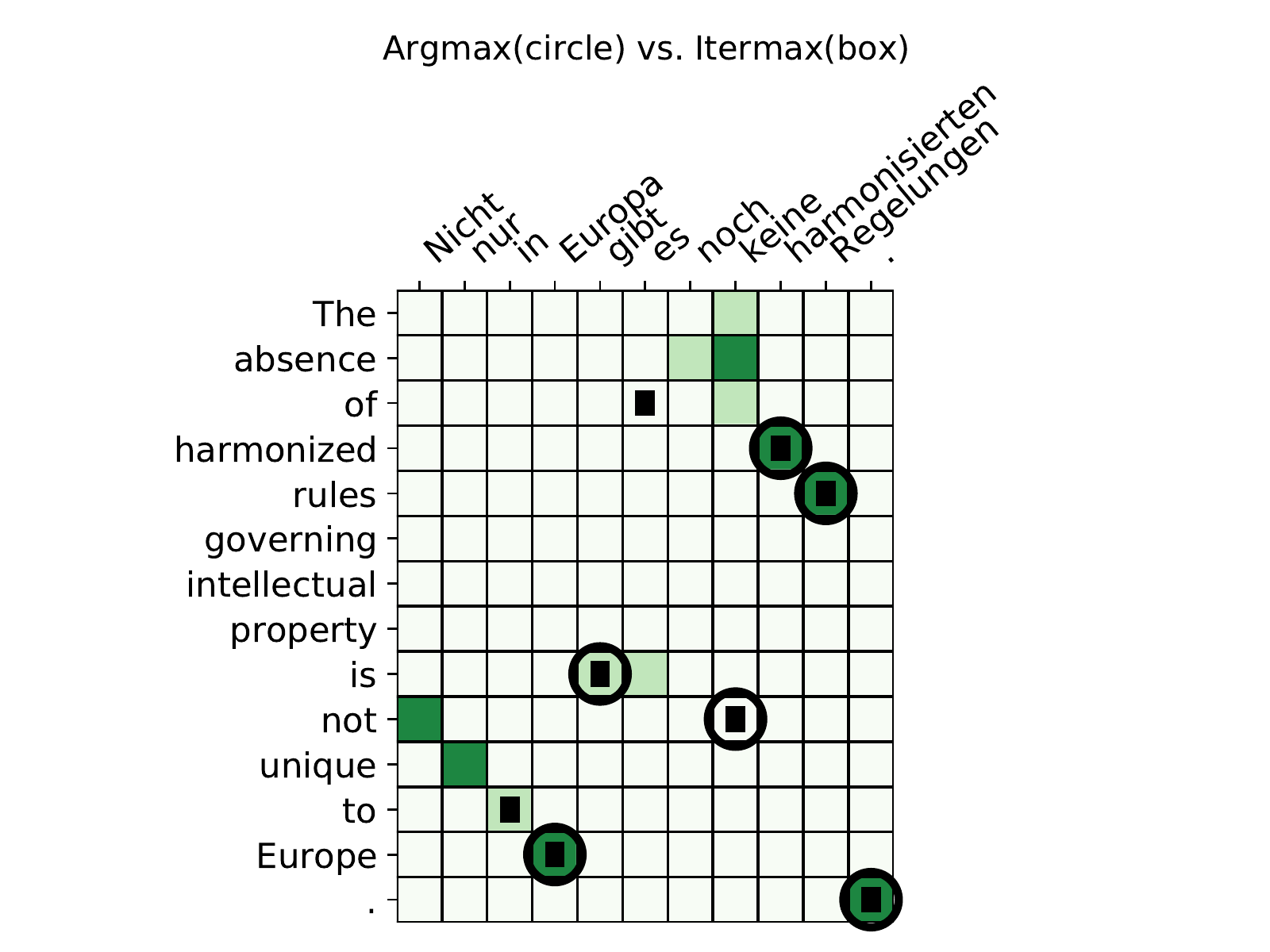}
	\includegraphics[width=0.95\linewidth]{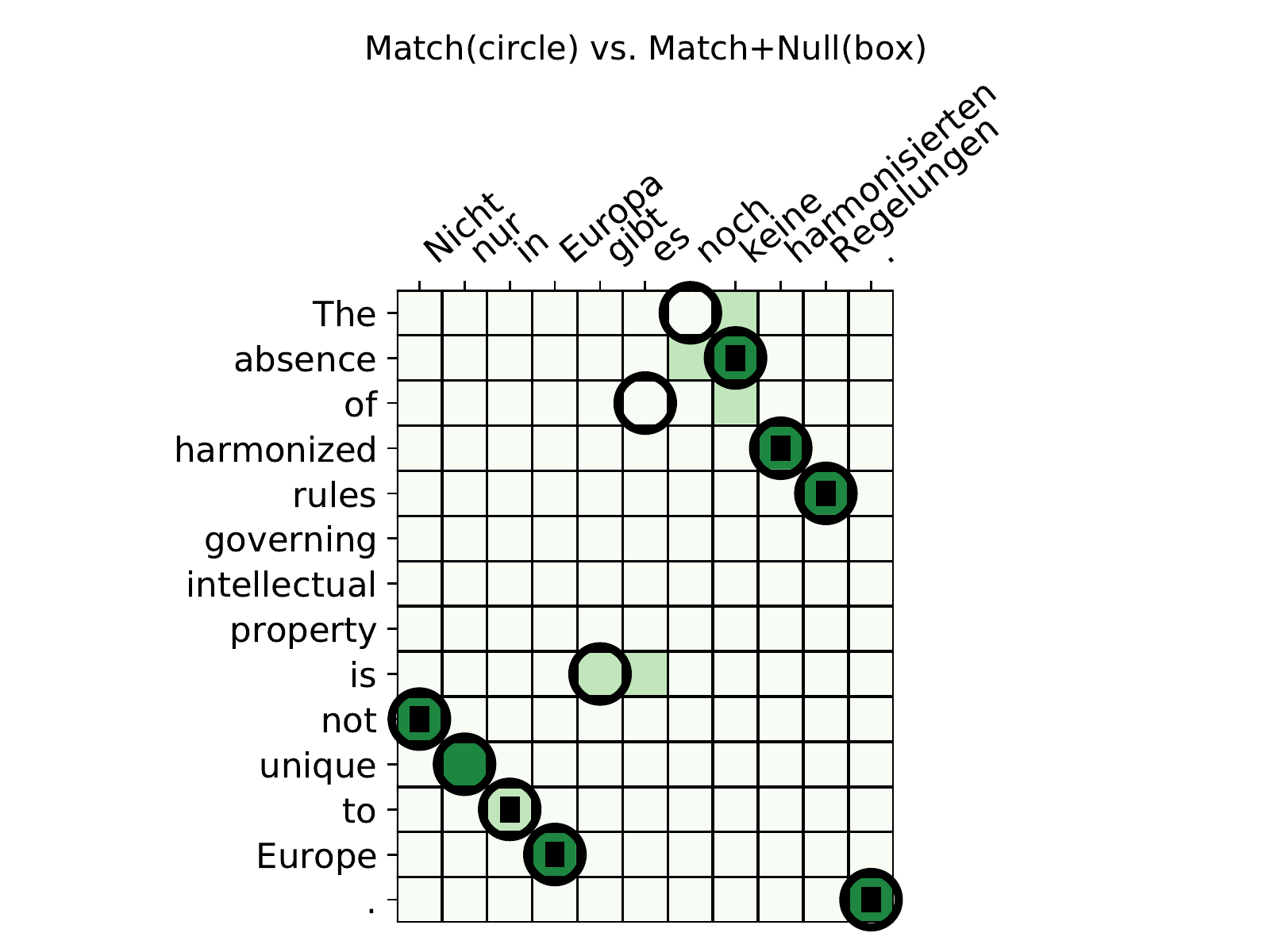}
	\includegraphics[width=0.95\linewidth]{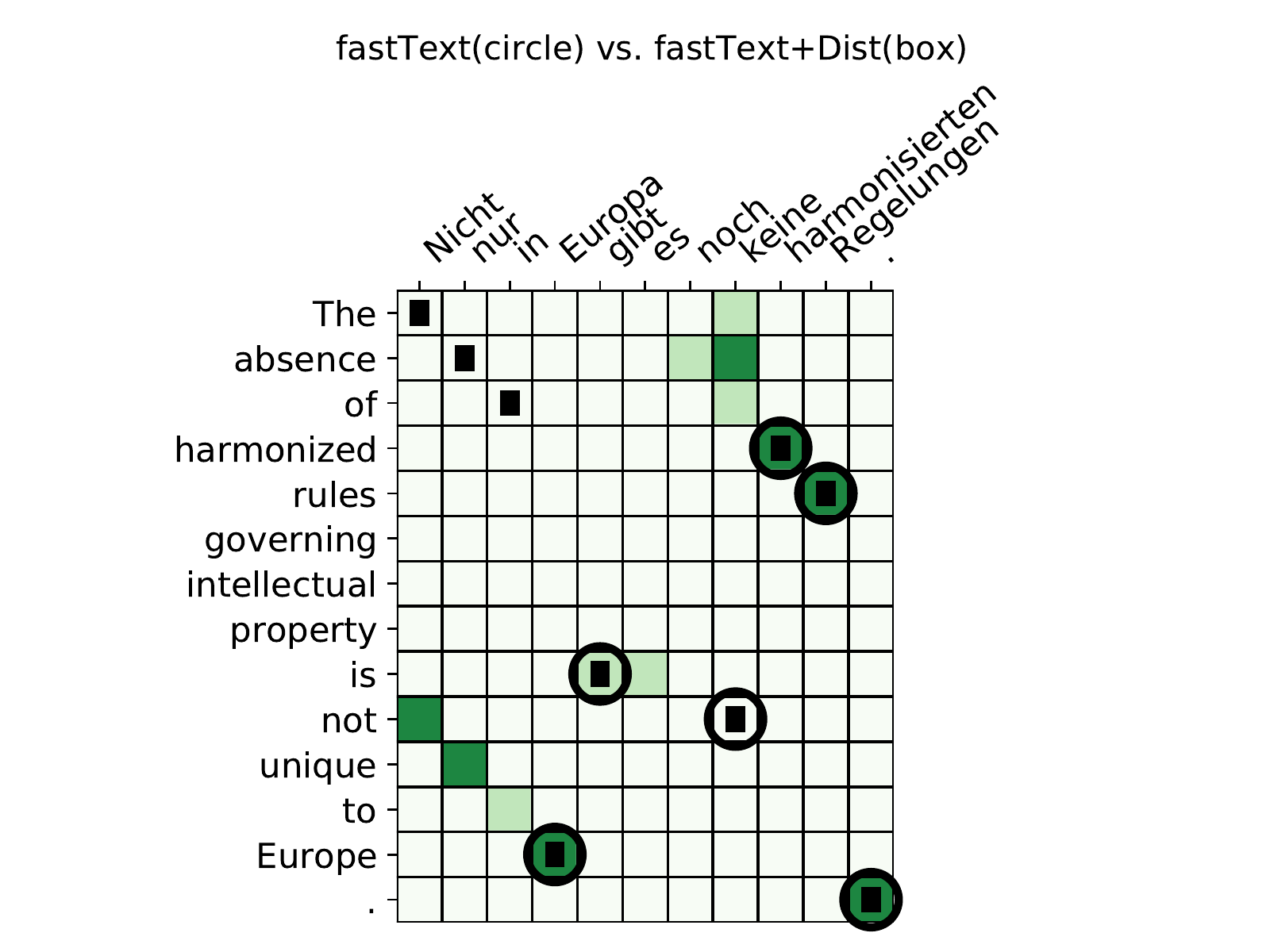}
	\caption{More examples.}
	\figlabel{examples3}
\end{figure}
\begin{figure}[t]
	\centering
	\includegraphics[width=0.95\linewidth]{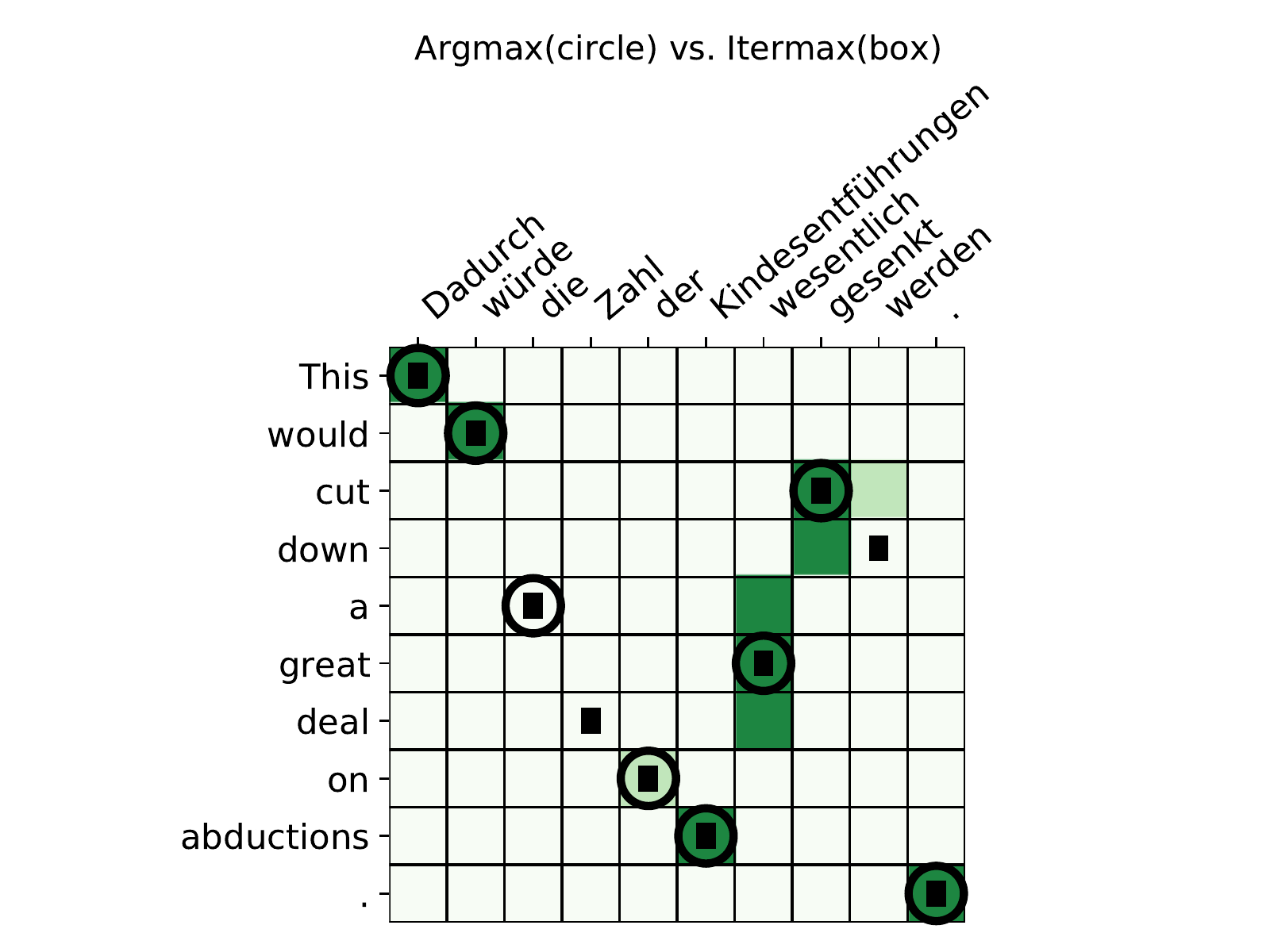}
	\includegraphics[width=0.95\linewidth]{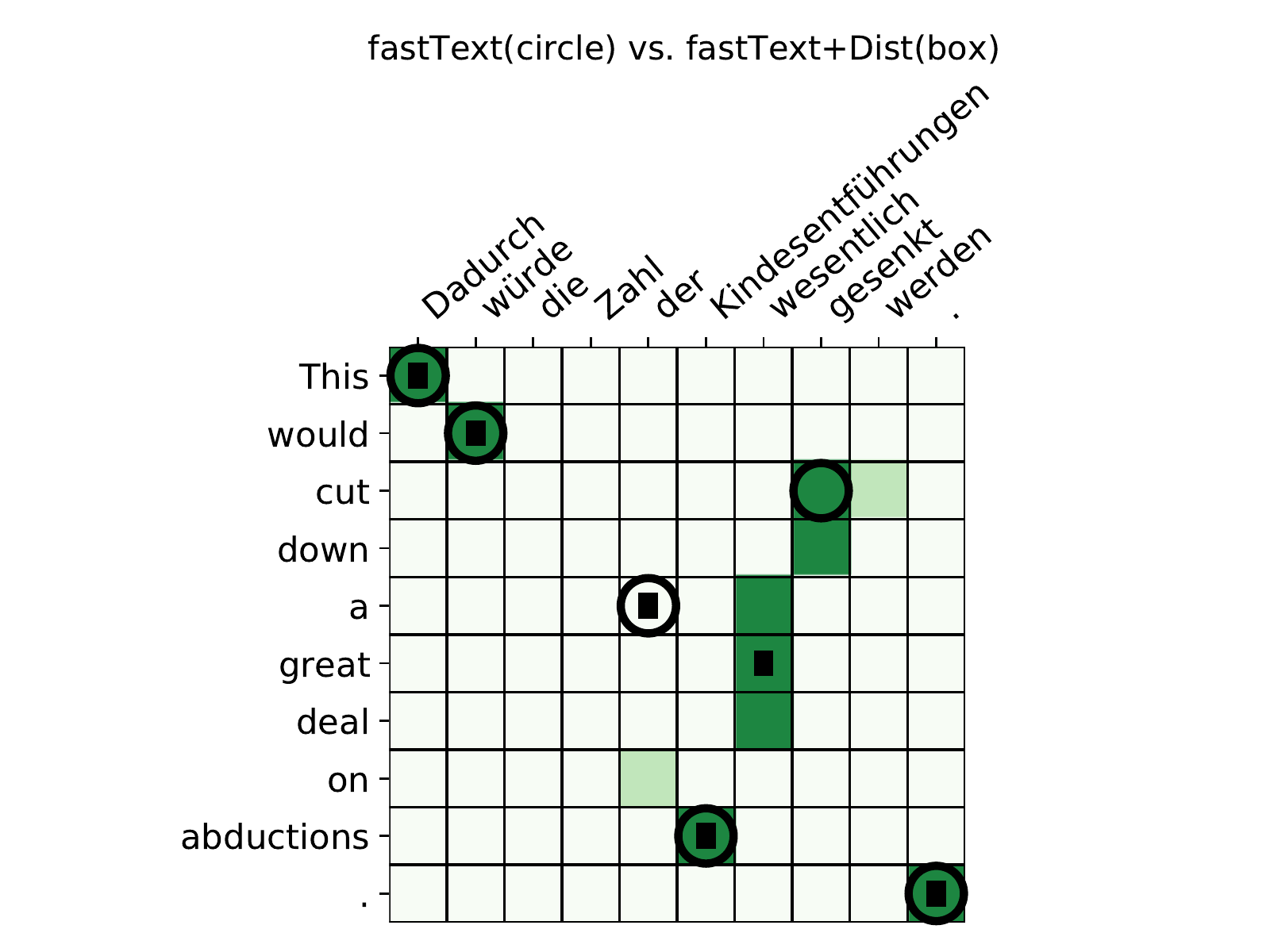}
	\includegraphics[width=0.95\linewidth]{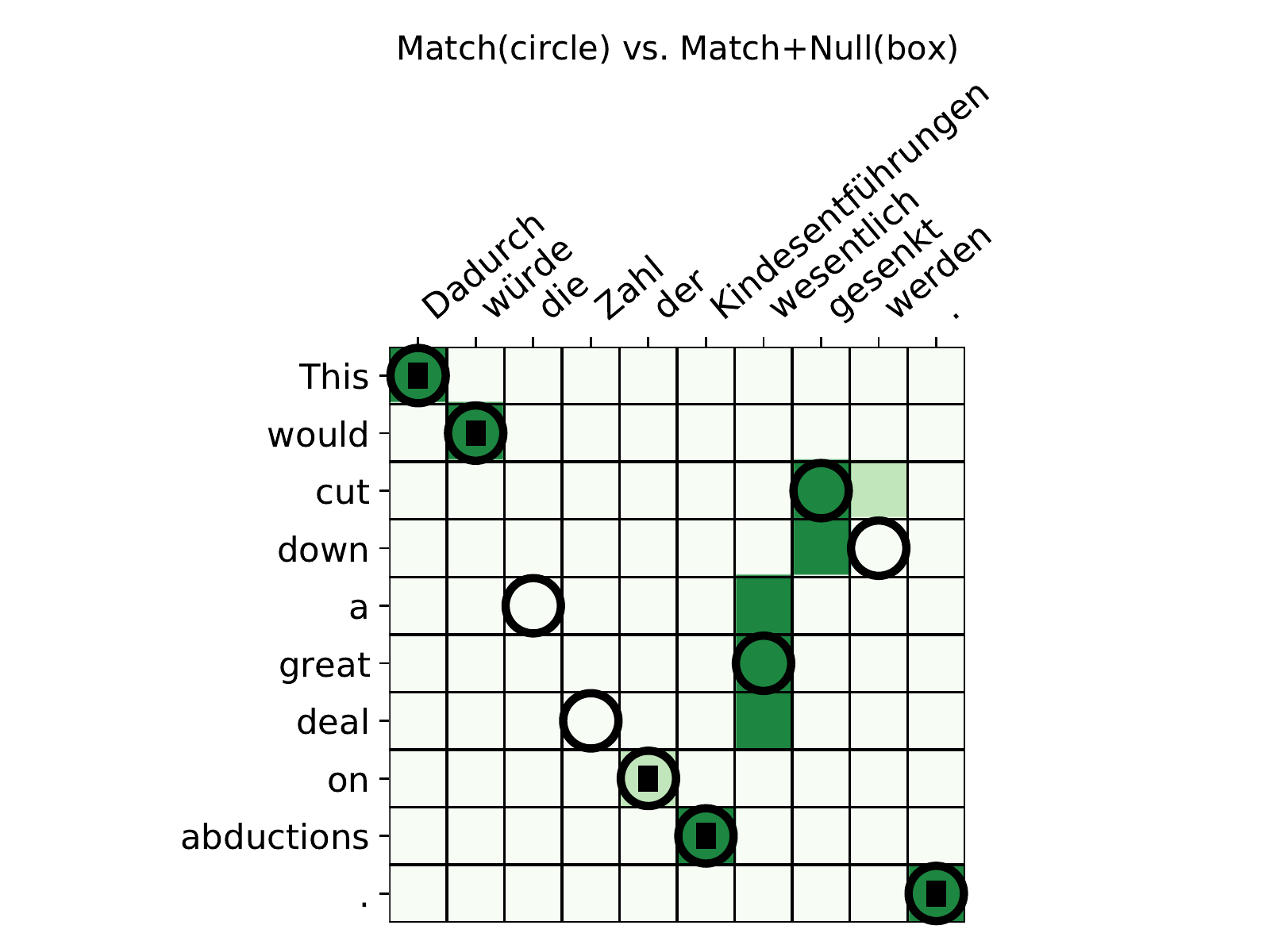}
	\caption{More examples.}
	\figlabel{examples4}
\end{figure}
\begin{figure}[t]
	\centering
	\includegraphics[width=0.95\linewidth]{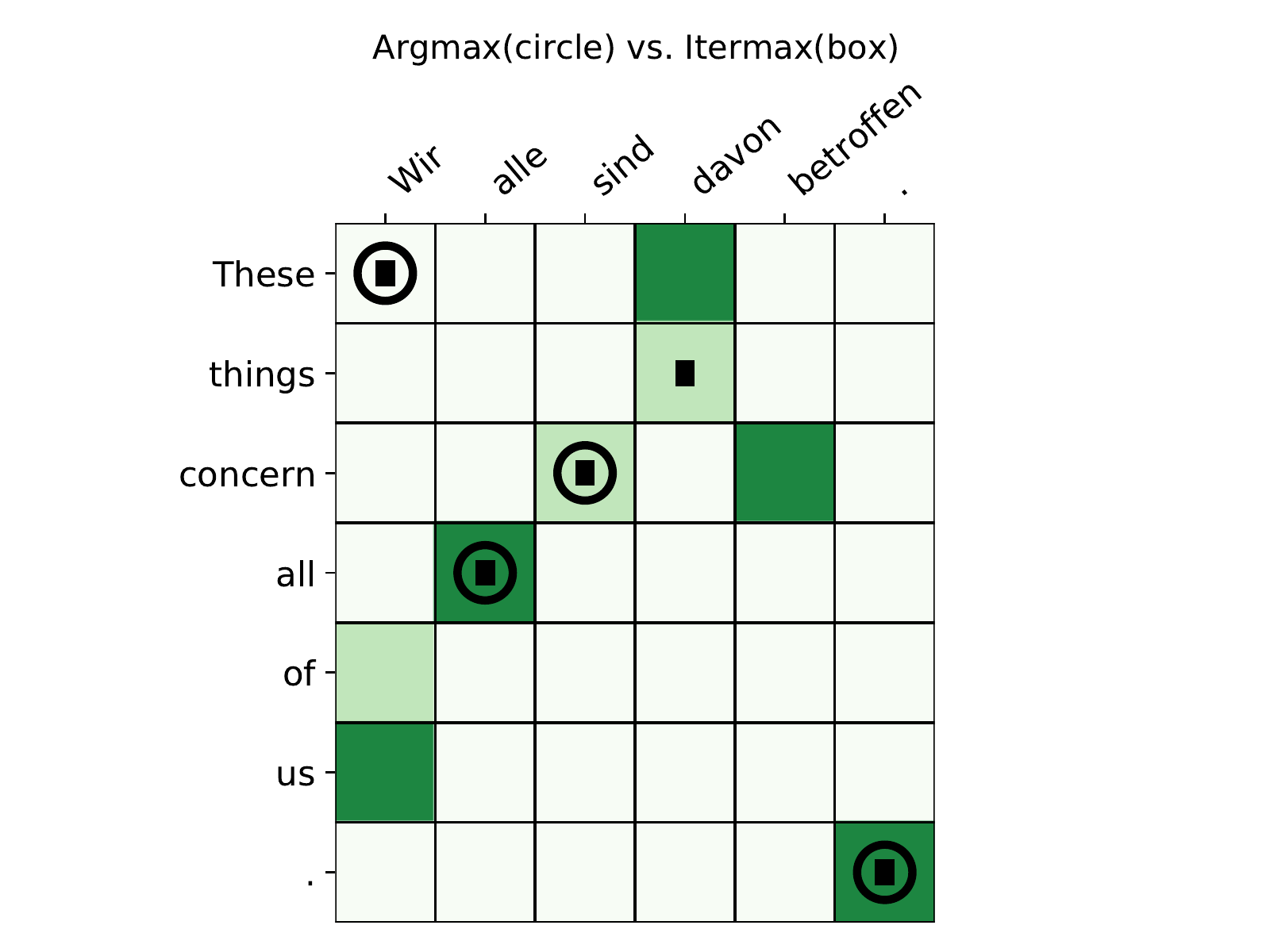}
	\includegraphics[width=0.95\linewidth]{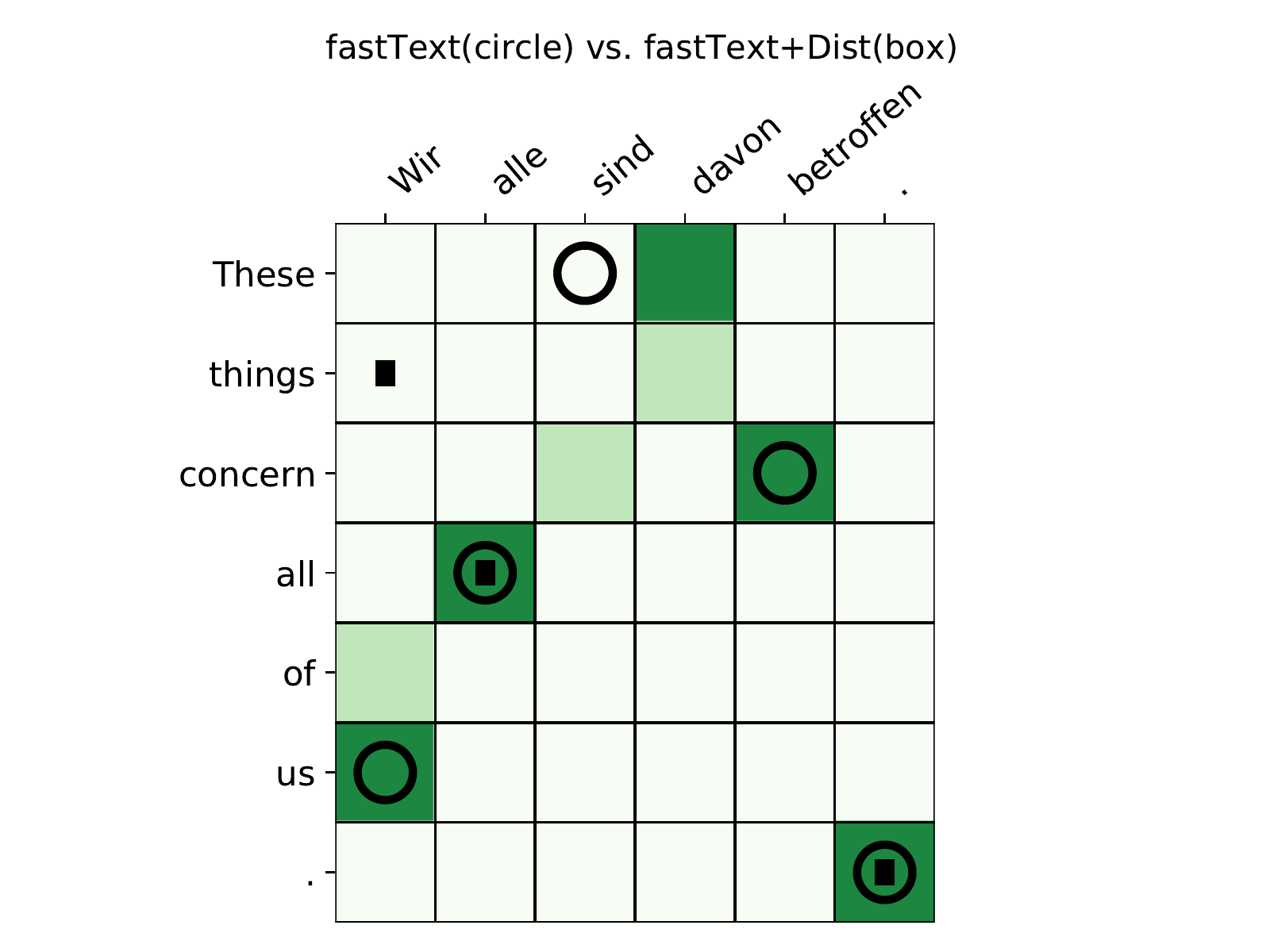}
	\includegraphics[width=0.95\linewidth]{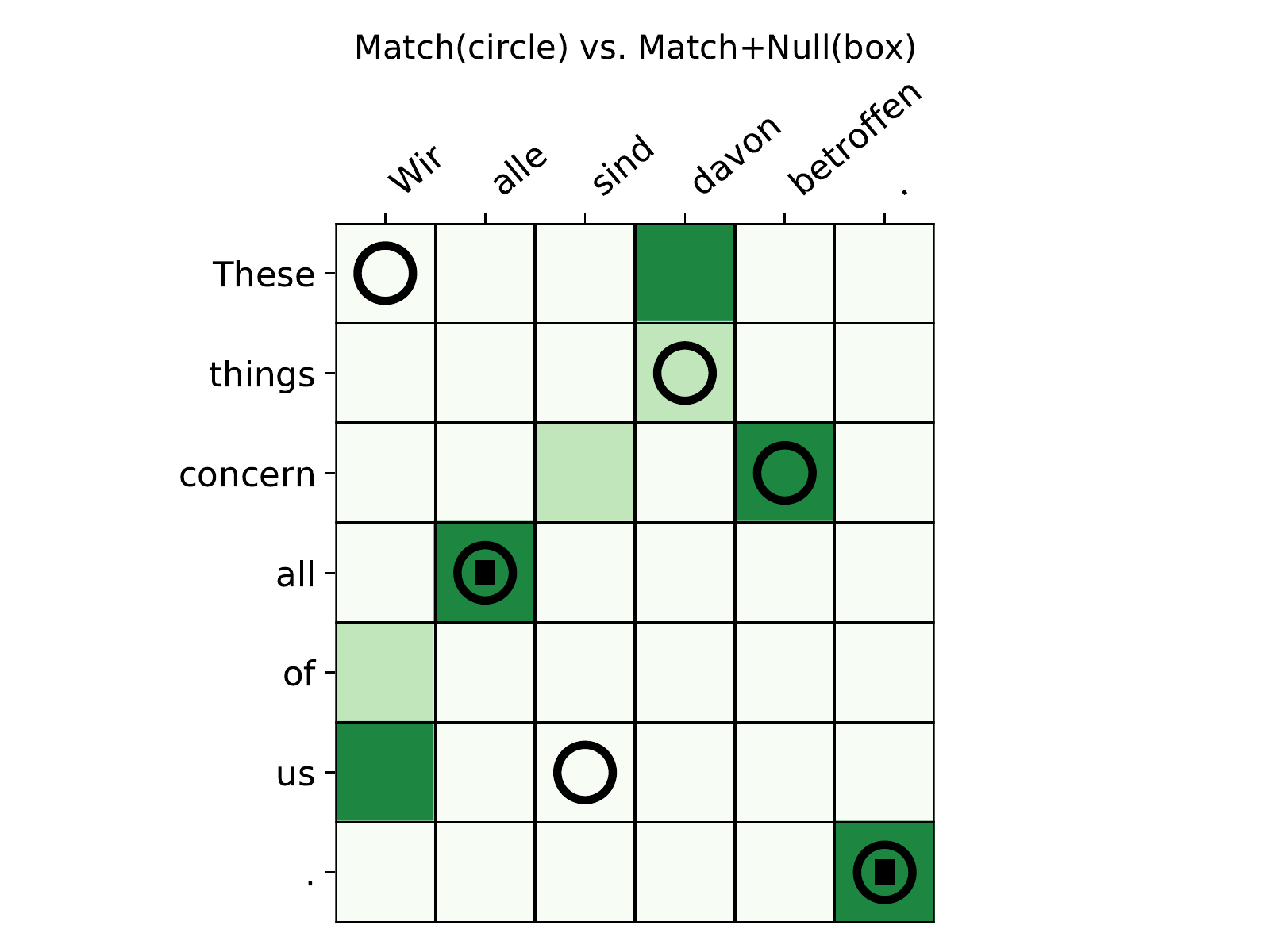}
	\caption{More examples.}
	\figlabel{examples5}
\end{figure}

\end{document}